\documentclass[preprint,12pt,authoryear]{elsarticle}
\usepackage[colorlinks, linkcolor=blue, filecolor=blue, urlcolor=blue, citecolor=cyan,bookmarks=false]{hyperref}
\usepackage{caption}
\usepackage[caption=false]{subfig}
\usepackage{graphicx}
\usepackage{multirow}
\usepackage{booktabs}
\usepackage{geometry}
\usepackage[noend]{algpseudocode}
\usepackage{algorithmicx,algorithm}
\usepackage{amssymb}
\usepackage{amsmath}
\usepackage{amsthm}
\usepackage{setspace}
\usepackage{lineno}
\usepackage{changepage}   % for the adjustwidth environment
\newcommand{\reconwidth}{0.13\textwidth}
\newcommand{\reffig}[1]{Fig.~\ref{#1}}
\newcommand{\refeq}[1]{Eq.~\eqref{#1}}
\begin{document}
\begin{frontmatter}
\title{
Bayesian imaging inverse problem with SA-Roundtrip prior via HMC-pCN sampler
}
\author[1]{Jiayu Qian}
\ead{jyqian@csu.edu.cn}

\author[1]{Yuanyuan Liu}
\ead{liuyy@csu.edu.cn}

\author[2]{Jingya Yang}
\ead{jingya_yang@tongji.edu.cn}

\author[1]{Qingping Zhou \texorpdfstring{\corref{cor1}}{}}
\cortext[cor1]{Corresponding author}
\ead{qpzhou@csu.edu.cn}

\affiliation[1]{organization={School of Mathematics and Statistics, Central South University}, city={Changsha}, postcode={Hunan 410083}, country={China}}
\affiliation[2]{organization={Shanghai Research Institute for Intelligent Autonomous Systems, Tongji University}, city={Shanghai}, postcode={Shanghai 200000}, country={China}}

\begin{abstract}
Bayesian inference with deep generative prior has received considerable interest for solving imaging inverse problems in many scientific and engineering fields. The selection of the prior distribution is learned from, and therefore an important representation learning of, available prior measurements. 
The SA-Roundtrip, a novel deep generative prior, is introduced to enable controlled sampling generation and identify the data's intrinsic dimension. This prior incorporates a self-attention structure within a bidirectional generative adversarial network. Subsequently, Bayesian inference is applied to the posterior distribution in the low-dimensional latent space using the Hamiltonian Monte Carlo with preconditioned Crank-Nicolson (HMC-pCN) algorithm, which is proven to be ergodic under specific conditions. Experiments conducted on computed tomography (CT) reconstruction with the MNIST and TomoPhantom datasets reveal that the proposed method outperforms state-of-the-art comparisons, consistently yielding a robust and superior point estimator along with precise uncertainty quantification.
\end{abstract}

%%Graphical abstract
% \begin{graphicalabstract}
% \begin{figure}[H]\centering
%     \includegraphics[height = 0.62\textheight]{Figure/F7.pdf}
%     \caption*{
%         Overall inversion algorithm diagram. STEP A: Use existing samples to train SA-Roundtrip. STEP B: Generators, sinogram, direct operator, and noise distributions are applied to HMC-pCN sampler to obtain the posterior distribution in latent space. STEP C: Samples are taken from a posterior distribution of latent space using a convergent Markov chain, and the samples are generated by a trained generator to infer the field. 
%     }
%      \label{fig:graph_abs}
% \end{figure}
% \end{graphicalabstract}

%%Research highlights
% \begin{highlights}
% \item We propose a new Bayesian framework for solving linear inverse problems. 

% \item We introduce a novel deep generative prior with the encoding and decoding structures.

% \item We verify the ergodicity of the HMC-pCN sampler under the proposed SA-Roundtrip prior.
% \end{highlights}


\end{frontmatter}

% \clearpage
\setcounter{page}{1}
% \pagewiselinenumbers
\section{Introduction}
\label{sec:Intro}
Imaging inverse problems are ubiquitous in various scientific and engineering fields, including denoising~\citep{houdard2018high,ashfahani2020devdan}, compressive sensing~\citep{bora2017compressed}, computed tomography (CT) reconstruction~\citep{baguer2020computed,hu2022dior,kofler2018u}.
These problems often involve inferring the model parameters of interest by combining the forward model with the constantly noisy observation. There has been a lot of research to solve this problem. Among them, Bayesian inference has received much attention recently due to its generalization and modularity~\citep{holden2021bayesian}, which offers much flexibility to deal with ill-posed inverse problems. In the Bayesian framework, the unknown images are treated as random variables. Then Bayes’ formula combines the prior information that encodes the information before observing the data and the observations modeled by their joint probability distribution conditioned on the parameters of interest to obtain a posterior distribution. A main challenge in Bayesian inference approaches here is to choose an accurate and informative prior from a collection of prior measurements.                                                                       
The standard approach has been to provide the prior analytically, called hand-crafted priors, chosen to encourage specific desired properties such as sparsity (e.g. $l^1$ prior~\citep{villena2009bayesian}), smoothness (e.g. Gaussian prior with an anisotropic Matern kernel~\citep{zhou2018approximate}), discontinuities (e.g. impulse prior and total variation prior~\citep{kaipio2007statistical}). 
While these hand-crafted priors capture some important aspects of the images, they are often overly simplistic and misspecified in the sense that they do not accurately describe their probability distribution, especially for high-dimensional image data.      
Alternative data-driven prior, known as deep generative models, embrace more flexibility, adaptivity, and effectiveness. They are designed to capture the statistics of the dataset. After training on a large dataset of clean images, these models  generate a complex distribution (e.g. that of natural images) from a simple latent base distribution (e.g. independent Gaussians) using a learned deterministic transformation. Once the deep generative model is successfully trained, its application to an inverse problem typically involves finding the optimal latent variable such that the resulting samples best fit the measurements. 
Existing methods for deep generative models in the literature can be broadly classified into four categories:
variational autoencoders~\citep{kingma2013auto,holden2021bayesian}, generative adversarial networks (GANs)~\citep{goodfellow2014generative,wu2019magnetic,lunz2018adversarial,patel2022solution}, normalizing flow~\citep{kingma2018glow,baso2022bayesian,zhou2019variational,whang2021solving}, and diffusion probabilistic model~\citep{ho2020denoising,cui2022self}. 
The first two approaches consist of generator networks that map a low-dimensional latent space to a high-dimensional image space. The latter two approaches learn a transformation network from a simple prior distribution to the target distribution. However, unlike the first two approaches, the prior distribution has the same dimension as the target distribution. 
While these models have achieved state-of-the-art performance in various image reconstruction tasks, recent works have highlighted their inability to quantify uncertainty. This is because many of these works are variational in nature, allowing them to obtain only a single point estimate~\citep{lunz2018adversarial,wu2019magnetic}. Additionally, there has been limited focus on identifying the latent space and understanding the theoretical properties of the resulting posterior distribution, such as the ergodicity of the sampling algorithm.                                              
This paper presents a solution within a comprehensive Bayesian framework. The method we introduce is a modified version of~\citet{patel2022solution}, which employs a GAN prior to address physics-based inverse problems. Our key enhancement to the GAN involves the addition of an encoding and decoding structure. This change allows us to precisely determine the uncertainty of the reconstruction by adjusting the intrinsic dimension of the $\boldsymbol{z}$ latent codes. On the other hand, the method outlined in~\citet{adler2018deep} also uses a GAN in a Bayesian context. However, their GAN model is trained to estimate the posterior distribution using paired samples of measurements and their corresponding true solutions.                                                   
\textcolor{black}{The benefits of the proposed method are confirmed through various numerical tests.} 
The primary contributions of this study are:
\begin{itemize}                        
    \item We introduce a novel SA-Roundtrip prior that involves a bidirectional generative adversarial network with self-attention structures.
    \item The development of a novel Bayesian framework designed to address linear inverse problems, which can detect the intrinsic dimension of the data by using the encoding and decoding structures of the SA-Roundtrip prior.
    \item A significant improvement compared to state-of-the-art Bayesian imaging techniques using GAN-based priors, as evidenced by various numerical experiments.
\end{itemize}

The rest of the paper is organized as follows. In Section~\ref{sec:setup}, we outline the problem setup for imaging inverse problems in the Bayesian framework. In Section~\ref{sec:prior}, we introduce the proposed deep generative prior, namely SA-Roundtrip prior. In Section~\ref{sec:method}, we present the Bayesian inference for the imaging inverse problems and prove that the Markov chain Monte Carlo sampling used is ergodic under certain conditions. In Section~\ref{sec:exp}, we present the implementation details and show the performed experiments on computed tomography reconstruction problems using two datasets. And in Section~\ref{sec:conclusion}, concluding remarks are presented.

\section{Problem setup}\label{sec:setup}
We consider an imaging problem of unknown image $\boldsymbol{x} \in \mathbb{R}^{d}$ and some clean observation data $\boldsymbol{y} \in \mathbb{C}^{p}$ and a statistical model with a likelihood function $p(\boldsymbol{y} \mid \boldsymbol{x})$ associates $\boldsymbol{x}$ with $\boldsymbol{y}$.
The mapping from $\boldsymbol{x}$ to $\boldsymbol{y}$ typically has a robust numerical solver. But what interests us is the ill-posed or ill-conditioned inverse issue of estimating $\boldsymbol{x}$ from $\boldsymbol{y}$ (i.e., the problem assumes that there is no unique solution that varies continuously with $\boldsymbol{y}$ or that there is a unique solution that is unstable with small perturbation).
Bayesian inference provides a way to solve inverse problems with quantified uncertainty estimates. \textcolor{black}{The unknown image and observation data} are modeled by random variables $\mathcal{X}$ and $\mathcal{Y}$, respectively. It is generally assumed that the observations are damaged by a noise $\boldsymbol{\eta}$, i.e., 
\begin{equation}\nonumber
    \hat{\boldsymbol{y}}=\boldsymbol{y}+\gamma  \boldsymbol{\eta},
\end{equation}
where $\gamma $ is the signal noise ratio. 
\textcolor{black}{Assume there is a prior distribution, denoted as $p_{\mathcal{X}}^{\text{prior}}$, which contains prior knowledge about the relationship between the observation variable $\mathcal{Y}$ and the unknown image variable $\mathcal{X}$.}
% Suppose we have a prior distribution $p_{\mathcal{X}}^{\text {prior }}(\boldsymbol{x})$ \textcolor{black}{over the unknown image $\boldsymbol{x}$}, which is derived from knowledge about $\mathcal{X}$ prior. 
Given $\mathcal{X}=\boldsymbol{x}$, we can construct a likelihood function $p_{\mathcal{Y}}^{\text{like}}(\hat{\boldsymbol{y}} \mid \boldsymbol{x})$ for the measurement $\hat{\boldsymbol{y}}$.
Applying the Bayesian rule, the posterior distribution of $\mathcal{X}$ is expressed as follows, 
\begin{equation}
p_{\mathcal{X}}^{\text{post}}(\boldsymbol{x} \mid \hat{\boldsymbol{y}})=\frac{p_{\mathcal{Y}}^{\text {like }}(\hat{\boldsymbol{y}} \mid \boldsymbol{x}) p_{\mathcal{X}}^{\text {prior }}(\boldsymbol{x})}{p_{\mathcal{Y}}(\hat{\boldsymbol{y}})} \propto p_{\eta}(\hat{\boldsymbol{y}}-\boldsymbol{f}(\boldsymbol{x})) p_{\mathcal{X}}^{\text{prior}}(\boldsymbol{x}),\label{eq:bayes_infer}
\end{equation}
where $p_{\mathcal{Y}}(\hat{\boldsymbol{y}})$ is called the evidence. In the likelihood term $p_{\mathcal{Y}}^{\text {like }}(\hat{\boldsymbol{y}} \mid \boldsymbol{x})=p_{\boldsymbol{\eta}}(\hat{\boldsymbol{y}}-f(\boldsymbol{x}))$, $\boldsymbol{x}$ is mapped to the same shape as $\boldsymbol{y}$ by direct operator $\boldsymbol{f}$. 

The Bayesian framework mentioned above has two significant challenges. First of all, it is difficult to create an acceptable prior expression $p_{\mathcal{X}}^{\text {prior }}(\boldsymbol{x})$ using previous knowledge learned from existing samples. For instance, it is typically impossible to represent images acquired as samples using standard distributions alone or in combination. Second, the posterior distribution for the majority of real-world inverse problems is situated in a high-dimensional space. As a result, it is impossible to determine its expectation using orthogonal-based or generic numerical techniques. In general, the Markov Chain Monte Carlo (MCMC) sampling method is a good solution to this issue. However, in this instance, the cost of utilizing this method to get a high-precision approximation of the expected value is rather expensive.

\section{The proposed SA-Roundtrip prior}\label{sec:prior}
In this section, we start by introducing the generative adversarial networks as well as Roundtrip, then presenting the proposed SA-Roundtrip prior. Specifically, some modifications are added in Section~\ref{sec:sa-roundtrip} to make this new deep generative prior applicable in practice.
\subsection{Generative Adversarial Networks}
Generative adversarial networks (GANs) are a class of generative models used for representation learning, was first proposed by Ian Goodfellow in~\citet{goodfellow2014generative}. 
The main idea of GANs is derived from the Nash equilibrium of game theory. The two parties involved in the game in GANs are a generator $\boldsymbol{G}$ and a discriminator $\boldsymbol{D}$ respectively. \textcolor{black}{The goal of the generator is to generate realistic samples by mapping a known distribution (e.g. normal, uniform, etc.) in a latent space to a distribution similar to the real data distribution;} the discriminator is a binary network that discriminates whether the input comes from the distribution of the real data or the generator. To win the game, these two players are required to continuously optimize, that is each improve their generative and discriminative abilities, and this learning optimization process is designed to find a Nash equilibrium between them.
Training these networks leads to the minimum maximum optimization of the loss function relative to the trainable parameter set. 
\textcolor{black}{Consider a random variable $\mathcal{Z}$ governed by a known distribution denoted by $p_\mathcal{Z}(\boldsymbol{z})$, where $\boldsymbol{z}\in \mathbb{R}^m$. The minimax objective for GANs can be formulated as follows:
\begin{equation}\label{eq:obj}
\underset{\boldsymbol{G}}{\min} \underset{\boldsymbol{D}}{\max}~~ \mathcal{L}_{GAN}(\boldsymbol{G},\boldsymbol{D}) = \mathbb{E}_{\boldsymbol{x}\sim p_{\mathcal{X}}(\boldsymbol{x})}[\log \boldsymbol{D}(\boldsymbol{x})] + \mathbb{E}_{\boldsymbol{z}\sim p_{\mathcal{Z}}(\boldsymbol{z})}[\log \boldsymbol{D}(\boldsymbol{\boldsymbol{G}(\boldsymbol{z})})].
\end{equation}}

Many GANs have been developed over the years~\citep{Radford2016UnsupervisedRL,8237566,8237506,brock2019large,9553597}. 
\textcolor{black}{One of the exemplary examples is self-attention generative adversarial networks (SAGAN)~\citep{9553597}.} SAGAN adds the self-attention block to GAN to improve the ability of $\boldsymbol{G}$ and discriminator $\boldsymbol{D}$ to model the global structure. Additionally, SAGAN employs spectral normalization in both $\boldsymbol{G}$ and $\boldsymbol{D}$, which not only ensures that the Lipschitz continuity condition is satisfied but also prevents gradient anomalies brought on by a large parameters in both models. Recently, WGAN-GP~\citep{gulrajani2017improved} has also been utilized as a prior for the Bayesian inverse problem~\citep{patel2022solution}.
% Therefore, we use WGAN-GP as a comparison model, and the results are shown in Section~\ref{sec:exp}.

\subsection{Roundtrip}
Roundtrip~\citep{Qiao2021Roundtrip} is a GAN framework with two discriminators ($\boldsymbol{D_z}$ and $\boldsymbol{D_x}$), an encoder $\boldsymbol{E}$, and a generator $\boldsymbol{G}$. $\boldsymbol{G}$ and $\boldsymbol{E}$ are used to learn the forward and backward mapping relationship distribution between them (STEP A of \reffig{fig:full_alg}), respectively.

\textcolor{black}{We denote $\tilde{\boldsymbol{x}}= \boldsymbol{G}(\boldsymbol{z})$ and $\tilde{\boldsymbol{z}}= \boldsymbol{E}(\boldsymbol{x})$}. The goal of $\boldsymbol{G}$ is to generate samples \textcolor{black}{$\tilde{\boldsymbol{x}}$} similar to the observed data \textcolor{black}{$\boldsymbol{x}$} while discriminator $\boldsymbol{D_x}$ is used to distinguish whether the data is observation data or generated data. 
\textcolor{black}{Similarly, $\boldsymbol{E}$ aims to generate samples $\tilde{\boldsymbol{z}}$ closely resembling the observed data $\boldsymbol{z}$, while discriminator $\boldsymbol{D_z}$ discriminates between the observational and encoded data.}
% encoder $\boldsymbol{E}$ and discriminator $\boldsymbol{D_z}$ attempt to map \textcolor{black}{$p_{\mathcal{X}}(\boldsymbol{x})$} to \textcolor{black}{$p_{\mathcal{Z}}(\boldsymbol{z})$.}
Furthermore, Roundtrip wants to minimize the distance between two data fields when a data point is transformed back and forth between them. This aims to guarantee that both $\boldsymbol{x} \rightarrow \boldsymbol{E(x)} \rightarrow \boldsymbol{G(E(x))}$ and $\boldsymbol{z} \rightarrow \boldsymbol{G(z)} \rightarrow \boldsymbol{E(G(z))}$ will remain in close proximity to the projection of $\boldsymbol{x}$ to the manifold caused by $\boldsymbol{G}$.

Utilizing a design that involves multiple generators and discriminators, Roundtrip exhibits the capability to establish mappings between two distributions in both directions. This characteristic translates into effective performance when applied to density estimation tasks. Nevertheless, the model structures of Roundtrip may not be suitable to represent complex prior distributions effectively (such as high-resolution images) when employed as priors within Bayesian inference, after getting undesired results.
 
\subsection{SA-Roundtrip}\label{sec:sa-roundtrip}
We modify the four networks ($\boldsymbol{G}$, $\boldsymbol{E}$, $\boldsymbol{D_{z}}$, and $\boldsymbol{D_{x}}$) of the original Roundtrip framework to improve its performance. Drawing inspiration from SAGAN, we adopt a residual network structure incorporating self-attention blocks and spectral normalization, to replace the conventional convolutional and fully-connected designs. Furthermore, we employ filter response normalization and threshold linear units, which respectively replace the partial batch normalization and partial rectified linear unit methods. Notably, these alternatives, as demonstrated in~\citet{9156322}, have proven to exhibit superior performance compared to other comparable techniques. A comprehensive delineation of the model architecture is offered in Section~\ref{sec:model_arch}.

\subsubsection{Self-attention mechanism}
The neural network receives input consisting of multiple vectors of varying sizes, each having certain relationships with one another. However, during actual training, these relationships among the inputs are not fully utilized, leading to unsatisfactory results in model training. This can be solved after introducing a self-attention mechanism. The self-attention mechanism actually wants the model to notice the correlation between different parts of the whole input so that it can take advantage of all the location information on the feature map, not just the local information. In addition, the discriminator with a self-attention mechanism can also check if the detailed features of distant parts of the image are consistent, which greatly improves the discriminant ability of the discriminator. 

Suppose there is an image feature $\boldsymbol{h}$ from the previously hidden layer, which we transform into two feature spaces $\boldsymbol{u}$, $\boldsymbol{\nu}$ to compute attention, where $\boldsymbol{u(h)}=\boldsymbol{W_{u}h}$, $\boldsymbol{\nu(h)}=\boldsymbol{W_{\nu}h}$.
\begin{equation}
\nonumber
\psi_{j, i}=\frac{\exp \left(s_{i j}\right)}{\sum_{i=1}^N \exp \left(s_{i j}\right)}, \text{where }s_{ij}=\boldsymbol{u}\left(\boldsymbol{h_{i}}\right)^T \boldsymbol{\nu}\left(\boldsymbol{h_j}\right),
\end{equation}
and $\psi_{j, i}$ denotes the attention of the model to the $i^{t h}$ location when forming the $j^{t h}$ region. Here, $C$ is the number of channels and $N$ is the number of feature locations of features from the previously hidden layer. The output of the attention layer is $\boldsymbol{o} = (\boldsymbol{o_1}, \boldsymbol{o_2}, \ldots, \boldsymbol{o_j}, \ldots, \boldsymbol{o_N}) \in$ $\mathbb{R}^{C \times N}$, where,
\begin{equation}
\nonumber
    \boldsymbol{o_j}=\boldsymbol{\phi}(\sum_{i=1}^N \psi_{j, i} \boldsymbol{g}(\boldsymbol{h_i})), \boldsymbol{g}(\boldsymbol{h_i})=\boldsymbol{W}_{\boldsymbol{g}} \boldsymbol{h_i}, \boldsymbol{\phi}(\boldsymbol{h_i})=\boldsymbol{W_\phi}\boldsymbol{h_{i}}.
\end{equation}
In the above formulation, $\boldsymbol{W_u} \in \mathbb{R}^{\bar{C} \times C}, \boldsymbol{W_{\nu}} \in \mathbb{R}^{\bar{C} \times C}$, $\boldsymbol{W_{g}} \in \mathbb{R}^{\bar{C} \times C}$, and $\boldsymbol{W_{\phi}} \in \mathbb{R}^{C \times \bar{C}}$ are the learned weight matrices, which are implemented as $1 \times 1$ convolutions and $\bar{C}=C / 8$, where $\bar{C}$ is the cropped channel, which can be set smaller according to the actual performance, so as to help reduce the computational cost.

In addition, we further multiply the output of the attention layer by a scale parameter and add back the input feature map. Therefore, the final output is given by,
\begin{equation}
\nonumber
\boldsymbol{h_i} = \lambda \boldsymbol{o_{i}} + \boldsymbol{h_{i}}, 
\end{equation}
where $\lambda$ is a learnable scalar and it is initialized as 0. A detailed introduction and theoretical derivation can be referred to~\citet{9553597}.

\subsubsection{SA-Roundtrip loss}
We will now introduce the loss function of SA-Roundtrip. The adversarial loss and the roundtrip loss may be separated into two categories. We employ the least squares loss function, which is compatible with that suggested in~\citet{8237566}, for the former.

The loss function during the training of the model can be expressed as:
\begin{equation}
\left\{\begin{aligned}
\mathcal{L}_{G A N}(\boldsymbol{G})=& \mathbb{E}_{\boldsymbol{z} \sim p_\mathcal{Z}(\boldsymbol{z})}\left[(\boldsymbol{D_{x}}(\boldsymbol{G}(\boldsymbol{z})-1)\right)^{2}] \\
\mathcal{L}_{G A N}\left(\boldsymbol{D_{x}}\right)=& \mathbb{E}_{\boldsymbol{x} \sim p_\mathcal{X}(\boldsymbol{x})}\left[(\boldsymbol{D_x} (\boldsymbol{x} )-1\right)^2]+\mathbb{E}_{\boldsymbol{z} \sim p_\mathcal{Z}(\boldsymbol{z})} [(\boldsymbol{D_x}(\boldsymbol{G}(\boldsymbol{z})))^2] \\
\mathcal{L}_{G A N}(\boldsymbol{E})=& \mathbb{E}_{\boldsymbol{x} \sim p_\mathcal{X}(\boldsymbol{x})}\left[(\boldsymbol{D_{z}}(\boldsymbol{E}(\boldsymbol{x}))-1\right)^{2}] \\
\mathcal{L}_{G A N}\left(\boldsymbol{D_{z}}\right)=& \mathbb{E}_{\boldsymbol{z} \sim p_\mathcal{Z}(\boldsymbol{z})}\left[(\boldsymbol{D_z}(\boldsymbol{z})-1\right)^{2}]+ \mathbb{E}_{\boldsymbol{x} \sim p_\mathcal{X}(\boldsymbol{x})} [(\boldsymbol{D_{z}}(\boldsymbol{E}(\boldsymbol{x})))^2]
\end{aligned}.\right.
\label{eq:2}
\end{equation}

Furthermore, in order to make the reconstruction results as close to the original data as possible, we choose to minimize the roundtrip loss with a cycle consistency loss function~\citet{8237506}. We define $\mathcal{D}_{\mathcal{Z}}(\boldsymbol{z}, \boldsymbol{E}(\boldsymbol{G}(\boldsymbol{z})))$, $\mathcal{D}_{\mathcal{X}}(\boldsymbol{x}, \boldsymbol{G}(\boldsymbol{E}(\boldsymbol{x})))$ denote the distance functions, which can be either $L_{1}$ or $L_{2}$ norm. 
Compared with the $L_{1}$ norm, the $L_{2}$ norm can prevent overfitting and improve the generalization ability of the model. Hence, the cycle consistency loss can be represented as
\begin{equation}\label{eq:3}
\begin{aligned}
    \mathcal{L}_{C C}(\boldsymbol{G}, \boldsymbol{E})=& \delta_{\boldsymbol{x}}\mathcal{D}_{\mathcal{X}}(\boldsymbol{x}, \boldsymbol{G}(\boldsymbol{E}(\boldsymbol{x}))) + \delta_{\boldsymbol{z}}\mathcal{D}_{\mathcal{Z}}(\boldsymbol{z}, \boldsymbol{E}(\boldsymbol{G}(\boldsymbol{z})))\\
    =&\delta_{\boldsymbol{x}}\|\boldsymbol{x}-\boldsymbol{G}(\boldsymbol{E}(\boldsymbol{x}))\|_{2}^{2}+ \delta_{\boldsymbol{z}}\|\boldsymbol{z}-\boldsymbol{E}(\boldsymbol{G}(\boldsymbol{z}))\|_{2}^{2},
    \end{aligned}
\end{equation}
where $\delta_{\boldsymbol{x}}$ and $\delta_{\boldsymbol{z}}$ represent the cycle consistency loss coefficients of $\boldsymbol{x}$ and $\boldsymbol{z}$, respectively. Considering both \refeq{eq:2} and (\ref{eq:3}), the total losses of the generator network and discriminator network are 
\begin{equation}\nonumber
\left\{\begin{aligned}
&\mathcal{L}(\boldsymbol{G}, \boldsymbol{E})=\mathcal{L}_{GAN}(\boldsymbol{G})+\mathcal{L}_{GAN}(\boldsymbol{E})+\mathcal{L}_{C C}(\boldsymbol{G}, \boldsymbol{E}) \\
&\mathcal{L}\left(\boldsymbol{D_{z}}, \boldsymbol{D_{x}}\right)=\mathcal{L}_{G A N}\left(\boldsymbol{D_{z}}\right)+\mathcal{L}_{G A N}\left(\boldsymbol{D_{x}}\right)
\end{aligned}.\right.
\label{eq:total}
\end{equation}

Joint training of two GAN models can be attempted by iteratively updating the parameters in two generators ($\boldsymbol{G}$ and $\boldsymbol{E}$) and two discriminators $\left(\boldsymbol{D_{z}}\right. $ and $\left. \boldsymbol{D_{x}}\right)$, respectively. Thus, differing from \refeq{eq:obj}, we represent the overall iterative optimization problem in SA-Roundtrip as
\begin{equation}\nonumber
\boldsymbol{G}^{*}, \boldsymbol{D^*_{x}}, \boldsymbol{E}^{*}, \boldsymbol{D^*_{z}}=\left\{\begin{array}{l}
\arg\underset{\boldsymbol{G}, \boldsymbol{E}}{\min} \ \mathcal{L}(\boldsymbol{G},\ \boldsymbol{E}) \\
\arg\underset{\boldsymbol{D_x}, \boldsymbol{D_z}}{\min} \ \mathcal{L}\left(\boldsymbol{D_z},\ \boldsymbol{D_x}\right)
\end{array}.\right.
\end{equation}
\begin{table}[!htb]
\caption{\\Comparison of the MMD, FID, and the number of trainable parameters. The downward arrow ($\downarrow$)  indicates that lower values are preferable.}
\centering
    \begin{tabular}{cccc}
    \toprule
        \hline \multicolumn{1}{c}{ Model } & \multicolumn{1}{c}{ \textcolor{black}{MMD ($\downarrow$)} } & \multicolumn{1}{c}{ FID ($\downarrow$) } & \multicolumn{1}{c}{ Parameters} \\
         \hline \multicolumn{1}{c}{ Roundtrip } & \multicolumn{1}{c}{ \textcolor{black}{0.088}} & \multicolumn{1}{c}{ 32 } & \multicolumn{1}{c}{8,665,921} \\
         \multicolumn{1}{c}{ WGAN-GP } &\multicolumn{1}{c}{ \textcolor{black}{0.049} } &  \multicolumn{1}{c}{ 20 } & \multicolumn{1}{c}{1,182,465} \\
         \multicolumn{1}{c}{ SA-Roundtrip } & \multicolumn{1}{c}{\textcolor{black}{0.034} } & \multicolumn{1}{c}{ 21 } & \multicolumn{1}{c}{1,276,257}\\
        \hline
        \bottomrule
        \end{tabular}\label{tb:three-prior-fid}
\end{table}

\noindent
\textbf{Remark 3.1.} 
As an example to demonstrate the capability of WGAN-GP, Roundtrip, and SA-Roundtrip priors, we perform them on the MNIST dataset and report \textcolor{black}{the maximum mean discrepancy(MMD)} as well as Fréchet Inception Distance(FID)~\citet{Martin2017fid} and the number of trainable parameters. \textcolor{black}{The MMD functions as a gauge of the disparity between the distributions of the two sample sets. Meanwhile, the FID quantifies the variety within the samples produced by the model.} The results of all aforementioned metrics are provided in \autoref{tb:three-prior-fid}.
        
\section{Bayesian inference for inverse problems}\label{sec:method}
\subsection{Posterior distribution in the latent variable}
We assume that the direct operator $f$ form in \refeq{eq:bayes_infer} is defined as follows:
\begin{equation}
\boldsymbol{f}(\boldsymbol{x})=\boldsymbol{Ax}, \boldsymbol{x} \in \mathbb{R}^d,
\label{eq:dir_ope}
\end{equation}
where $\boldsymbol{A}$ is the observation operator $\boldsymbol{A} \in \mathbb{C}^{d \times p}$ and it is rank deficient, or problems where $\boldsymbol{A}^{\top} \boldsymbol{A}$ is full rank but has a poor condition number. 

%GAN的本质就是通过极大似然估计方法去学习隐变量z到真实数据分布的映射，利用GAN作为贝叶斯推断的先验的合理性已经在"BAYESIAN IMAGING WITH DATA-DRIVEN PRIORS ENCODED BY NEURAL NETWORKS: THEORY, METHODS, AND ALGORITHMS"的第3.1节中已有说明。并且，SA-Roundtrip中所用到的是最小二乘损失，因此该模型理论上是LSGAN的变种，优化这个损失就是最小化皮尔森卡方散度，xx等人在"Least Squares Generative Adversarial Networks"中对其收敛性做了详细推导。
\textcolor{black}{
For a generator $\boldsymbol{G}^*$ that is perfectly trained using samples from the true distribution $p_{\mathcal{X}}^{\text{true}}$, the generator distribution $p_{\mathcal{X}}^{\text{g}}$ should be equal to the true distribution. We set the prior distribution to match the true distribution, that is $p_{\mathcal{X}}^{\text{prior}} = p_{\mathcal{X}}^{\text{true}} = p_{\mathcal{X}}^{\text{g}}$. 
Therefore,
\begin{equation}\nonumber
    p_{\mathcal{X}}^{\text {post }}(\boldsymbol{x} \mid \hat{\boldsymbol{y}})\propto p_{\eta}(\hat{\boldsymbol{y}}-\boldsymbol{f}(\boldsymbol{x})) p_{\mathcal{X}}^{\text {g}}(\boldsymbol{x}).
\end{equation}
For a sufficiently smooth function $l$, we have
\begin{equation}\label{empirically}
    \underset{\boldsymbol{x} \sim p_{\mathcal{X}}^{\text {post }}}{\mathbb{E}}[l(\boldsymbol{x})] = \underset{\boldsymbol{z} \sim p_{\mathcal{Z}}^{\text {post }}}{\mathbb{E}}[l(\boldsymbol{G}^*(\boldsymbol{z}))],
\end{equation}
Empirically, this equation is observed in a perfectly trained generative model. Similar results are also found in the works of \citet{holden2021bayesian}, \citet{gonzalez2022solving}, and \citet{pandit2020inference}.
This leads to:
\begin{equation}\nonumber
    p_{\mathcal{Z}}^{\text {post }}(\boldsymbol{z} \mid \hat{\boldsymbol{y}})\propto p_{\eta}(\hat{\boldsymbol{y}}-\boldsymbol{f}(\boldsymbol{G}^*(\boldsymbol{z}))) p_{\mathcal{Z}}(\boldsymbol{z}).
\end{equation}
The above equation implies that sampling from the posterior distribution for $\boldsymbol{x}$ is equivalent to sampling from the posterior distribution for $\boldsymbol{z}$ and then
transforming the sample through the generator $\boldsymbol{G}^*$. In other words,
\begin{equation}\nonumber
\boldsymbol{x} \sim p_\mathcal{X}^{\text{post}}(\boldsymbol{x}|\hat{\boldsymbol{y}}) \Rightarrow \tilde{\boldsymbol{x}} = \boldsymbol{G}^*(\boldsymbol{z}), \boldsymbol{z} \sim p_\mathcal{Z}^{\text{post}}(\boldsymbol{z}|\hat{\boldsymbol{y}}).
\end{equation}}
% \textcolor{black}{
% A lot of literature uses a similar approach. Although we do not prove it theoretically, empirically Eq. (\ref{empirically}) still holds, e.g., \citep{ holden2021bayesian}
% }
In the above setting, we reformulate the direct operator $\boldsymbol{f}$ as follows:
\begin{equation}
\nonumber
\begin{split}
    \boldsymbol{f}(\boldsymbol{z})&=\boldsymbol{A}\boldsymbol{G}^{*}(\boldsymbol{z}), \boldsymbol{z} \in \mathbb{R}^m.
    \end{split}
\end{equation}

Consider the example $\hat{\boldsymbol{y}}=\boldsymbol{A}\boldsymbol{x}+\boldsymbol{\eta}$, where $\boldsymbol{\eta} \sim \mathcal{N}\left(0,\sigma^{2} \mathbb{I}_p\right)$ and $\boldsymbol{G}^{*}$ as a prior distribution of Bayesian inference and here $p_{\mathcal{Z}}^{\text {post }}$ is the posterior distribution of $\boldsymbol{z}$ given $\hat{\boldsymbol{y}}$ and is defined as, 
\begin{equation}
\begin{aligned}
p_{\mathcal{Z}}^{\text {post }}(\boldsymbol{z} \mid \hat{\boldsymbol{y}}) &= \frac{p_{\boldsymbol{\eta}}\left(\hat{\boldsymbol{y}}-\boldsymbol{f}\left(\boldsymbol{G}^{*}(\boldsymbol{z})\right)\right) p_{\mathcal{Z}}(\boldsymbol{z})}{p_{\mathcal{Y}}(\hat{\boldsymbol{y}})}\\
&\propto p_{\boldsymbol{\eta}}(\hat{\boldsymbol{y}}-\boldsymbol{f}(\boldsymbol{G}^{*}(\boldsymbol{z}))) p_{\mathcal{Z}}(\boldsymbol{z})\\
&\propto \exp \left\{-\frac{\left(\hat{\boldsymbol{y}}-\boldsymbol{A} \boldsymbol{G}^{*}(\boldsymbol{z})\right)^{\dagger}\left(\hat{\boldsymbol{y}}-\boldsymbol{A} \boldsymbol{G}^{*}(\boldsymbol{z})\right)}{2 \sigma^2}-\frac{\boldsymbol{z}^{\top} \boldsymbol{z}}{2}\right\}.
\end{aligned}
\label{eq:post_dis}
\end{equation}

We can evaluate the statistics on $p_{\mathcal{X}}^{\text {post }}$ by sampling $\boldsymbol{z}$ from $p_{\mathcal{Z}}^{\text{post}}$ and passing the sample to $\boldsymbol{G}^{*}.$ Since the expression of $p_{\mathcal{Z}}^{\text {post }}$ is known and $m \ll d$, we can use MCMC algorithm to sample from $p_{\mathcal{Z}}^{\text {post }}$ efficiently. It is worth noting that by reformulating the high-dimensional posterior with respect to image space $\boldsymbol{x}$ to the low-dimensional latent space $\boldsymbol{z}$, we can efficiently explore the posterior distribution using MCMC algorithms, which will be described in the next section.

\subsection{Markov Chain Monte Carlo}In this section, based on the preconditioned Crank-Nicolson (pCN) algorithm and Hamilton Monte Carlo (HMC) sampler, we present the proposed HMC-pCN algorithm. 

\subsubsection{Preconditioned Crank-Nicolson algorithm}
The preconditioned Crank-Nicolson (pCN)~\citep{Cotter2013MCMC} algorithm was originally introduced as the progressive iterative approximation algorithm in~\citet{BESKOS2008MCMC}. 
For random walk Metropolis and traditional gradient-based proposals, the acceptance probability tends to zero as the dimensionality increases. In contrast, the acceptance probability and convergence properties of pCN are robust to increasing dimensionality~\citep{Martin2014Spectral}, making pCN useful for high-dimensional models.
Its key idea is that using the Ornstein-Uhlenbeck proposal
\begin{equation}
\boldsymbol{z}_{t}= \sqrt{1-\beta^{2}} \boldsymbol{z}_{t-1}+\beta \boldsymbol{\xi}_{t}.
\label{eq:pcn}
\end{equation}
This equation can greatly improve the performance of the standard Metropolis-Hastings algorithm when targeting a measure that is a reweighting of a Gaussian.
The proposal is a first-order autoregressive process, rather than a centered random walk. 

The associated acceptance probability is given by
\begin{equation}
\nonumber
\alpha(\boldsymbol{z}, \boldsymbol{z}^\ast)=\min \{1, \exp (\Phi(\boldsymbol{z})-\Phi(\boldsymbol{z}^\ast))\},
\end{equation} 
where $\Phi(\boldsymbol{z})={\|\left(\hat{\boldsymbol{y}}-\boldsymbol{A} \boldsymbol{G}^{*}(\boldsymbol{z})\right)\|_{2}^{2}}/{2 \sigma^2} + \|\boldsymbol{z}\|^2_2/2$.

\subsubsection{Hamiltonian Monte Carlo with pCN}
Hamilton Monte Carlo (HMC)~\citep{DUANE1987216} is an efficient gradient-based sampling algorithm.
The HMC algorithm consists of two steps: 1) the molecular dynamics step and 2) the Monte Carlo step.
The molecular dynamics step involves integrating Hamiltonian dynamics. The Monte Carlo step uses Metropolis-Hastings to deal with any errors introduced by the molecular dynamics step when using numerical integrators~\citep{neal2011hmcm}. 

In HMC, the position vector $z$ is augmented by an auxiliary momentum variable $\boldsymbol{r}$, which is usually considered to be independent of $\boldsymbol{z}$. The Hamiltonian $H (\boldsymbol{z}, \boldsymbol{r})$ representing the total energy of the system is as follows:
\begin{equation}\label{eq:hami}
    H(\boldsymbol{z}, \boldsymbol{r})=U(\boldsymbol{z})+K(\boldsymbol{r}),
\end{equation}
where $U(\boldsymbol{z})$ is potential energy or the negative log-likelihood of the target posterior distribution, and its expression is as follows:
\begin{equation}
    U(\boldsymbol{z})=\frac{\|\left(\hat{\boldsymbol{y}}-\boldsymbol{A} \boldsymbol{G}^{*}(\boldsymbol{z})\right)\|_{2}^{2}}{2 \sigma^2}+\frac{\|\boldsymbol{z}\|_{2}^{2}}{2}.
\end{equation}

$K(\boldsymbol{r})$ represents the kinetic energy derived from the Gaussian kernel with a covariance matrix $\mathcal{M}$.
\begin{equation}
K(\boldsymbol{r})=\frac{1}{2}\log((2\pi)^d|\mathcal{M}|)+\frac{1}{2}\boldsymbol{r}^T\mathcal{M}\boldsymbol{r}.
\label{eq:kinetic}
\end{equation}

Since the Hamiltonian in \refeq{eq:hami} is separable, the most commonly used numerical integration scheme in HMC is the leapfrog. The equation for using the leapfrog to update position and momentum is as follows:
\begin{equation}\label{eq:leapfrog}
\begin{aligned}
& \boldsymbol{r}_{t+\frac{\epsilon}{2}}=\boldsymbol{r}_t+\frac{\epsilon}{2} \frac{\partial H\left(\boldsymbol{z}_t, \boldsymbol{r}_t\right)}{\partial \boldsymbol{z}}, \\
& \boldsymbol{z}_{t+\epsilon}=\boldsymbol{z}_t+\epsilon \mathcal{M}^{-1} \boldsymbol{r}_{t+\frac{\epsilon}{2}}, \\
& \boldsymbol{r}_{t+\epsilon}=\boldsymbol{r}_{t+\frac{\epsilon}{2}}+\frac{\epsilon}{2} \frac{\partial H\left(\boldsymbol{z}_{t+\epsilon}, \boldsymbol{r}_{t+\frac{\epsilon}{2}}\right)}{\partial \boldsymbol{z}},
\end{aligned}
\end{equation}
where $\epsilon$ is the discretization step size.

In order to adequately sample from the model, the momentum needs to be re-sampled from the normal distribution at each iteration. 
We assume that the model being trained is a continuous space, and if this model is well trained (i.e., low FID enough), then similar samples in this space should be adjacent to each other. If the two re-samplings of momentum are uncorrelated(e.g., fully re-sample the moment), then the two results may be quite different. 

We propose to use \refeq{eq:pcn} instead of the full re-sampling step in HMC. This approach will cause the proposed momentum at each step to retain the information of the previous momentum. The expression for the partial momentum refreshment using pCN is as follows: 
\begin{equation}
\boldsymbol{r}_{t}= \sqrt{1-\beta^{2}} \boldsymbol{r}_{t-1}+\beta \boldsymbol{\xi}_{t},\ \boldsymbol{\xi}_{t} \sim N(0,\ \mathcal{M}), 
\label{eq:resampling}
\end{equation}
where $\beta$ is an artificially specified refresh parameter that takes values between 0 and 1. When $\beta$ is equal to one, the momentum is never updated and when $\beta$ is equal to zero, the momentum is always updated. 

Since discretization errors arise during numerical integration, the Monte Carlo step in HMC utilizes the Metropolis-Hastings to make the parameters $\boldsymbol{z}^{*}$ and $\boldsymbol{r}^{*}$ proposed by its molecular dynamics step accepted with probability:
\begin{equation}
\alpha\left(\boldsymbol{z}^{*},\boldsymbol{r}^{*}\right)=\min \{1, \exp (H(\boldsymbol{z},\ \boldsymbol{r})-H\left(\boldsymbol{z}^{*},\ \boldsymbol{r}^{*}\right))\}.
\label{eq:acp}
\end{equation}

Algorithm \ref{alg:hmc} and algorithm \ref{alg:hmc_pcn} show the HMC and HMC-pCN samplers, respectively. 

\begin{algorithm}[!htb]
    \caption{Hamilton Monte Carlo}
    \hspace*{0.02in} {\bf Input:} 
    Initial position $\boldsymbol{z}_{(0)}$ and discretization step size $\epsilon$.
    \begin{algorithmic}[1]
    \For{$t=1,\ 2,\ . . . $} 
    \State $\boldsymbol{r}_{(t-1)} \sim N(0,\ \mathcal{M})$
    \State $\boldsymbol{z}^{*}, \boldsymbol{r}^{*} = \textbf{Leapfrog}(\boldsymbol{z}_{(t-1)}, \boldsymbol{r}_{(t-1)}, \epsilon)$ in~\refeq{eq:leapfrog}
    \State Set $\boldsymbol{z}_{(t)}=\boldsymbol{z}^{*}$ with probability $\alpha$ in ~\refeq{eq:acp}
    \State Set $\boldsymbol{z}_{(t)}=\boldsymbol{z}_{(t-1)}$ otherwise. 

    \EndFor
    
    \State \Return $\{\boldsymbol{z}_{(1)},\ \boldsymbol{z}_{(2)},\ . . .\ ,\ \boldsymbol{z}_{(t)}\}$ 
    
    \end{algorithmic}
    \label{alg:hmc}
    \end{algorithm}
\begin{algorithm}[!htb]
    \caption{Hamilton Monte Carlo with pCN}
    \hspace*{0.02in} {\bf Input:} 
    Initial position $\boldsymbol{z}_{(0)}$, initial momentum $\boldsymbol{r}_{(0)}$ and discretization step size $\epsilon$.
    \begin{algorithmic}[1]
    \For{$t=1,\ 2,\ . . . $} 
    \State $\boldsymbol{z}^{*}, \boldsymbol{r}^{*} = \textbf{Leapfrog}(\boldsymbol{z}_{(t-1)}, \boldsymbol{r}_{(t-1)}, \epsilon)$ in~\refeq{eq:leapfrog}
    \State Set $\boldsymbol{z}_{(t)}=\boldsymbol{z}^{*}$ and $\boldsymbol{r}_{(t)}=\boldsymbol{r}^{*}$ with probability $\alpha$ in~\refeq{eq:acp}
    \State Set $\boldsymbol{z}_{(t)}=\boldsymbol{z}_{(t-1)}$ and $\boldsymbol{r}_{(t)}=\boldsymbol{r}_{(t-1)}$ otherwise. 
    \State $\boldsymbol{r}_{(t)}=\sqrt{1-\beta^{2}} \boldsymbol{r}_{(t)}+\beta \boldsymbol{\xi}_{(t)},\ \boldsymbol{\xi}_{(t)} \sim N(0,\ \mathcal{M})$

    \EndFor
    
    \State \Return $\{\boldsymbol{z}_{(1)},\ \boldsymbol{z}_{(2)},\ . . .\ ,\ \boldsymbol{z}_{(t)}\}$ 
    
    \end{algorithmic}
    \label{alg:hmc_pcn}
    \end{algorithm}
\subsection{Ergodicity of Markov Chain}
The Markov chain produced by the sampling approach is designed to accommodate $p_{\mathcal{Z}}^{\text {post }}(\boldsymbol{z} \mid \hat{\boldsymbol{y}})$ as an invariant density. However, we also need to verify that the Markov chain is ergodic in order to guarantee convergence to $p_{\mathcal{Z}}^{\text {post }}(\boldsymbol{z} \mid \hat{\boldsymbol{y}})$ and ensure that the samples generated may be utilized for Monte Carlo estimations of expectation. It is sufficient to demonstrate that the following criteria are met~\citet[Assumptions 6.1 and Theorem 6.2]{Cotter2013MCMC} for this to hold.

\textbf{Conditions 4.1.}
\emph{
Suppose $\mathcal{F}$ is given by the potential of the likelihood}
\begin{equation}
\nonumber
\mathcal{F}(\boldsymbol{z} ; \hat{\boldsymbol{y}})=\frac{\left\|\hat{\boldsymbol{y}}-A \boldsymbol{G}^{*}(\boldsymbol{z})\right\|^2}{2 \sigma^2},
\end{equation}

\emph{then the function $\mathcal{F}: \mathcal{Z} \rightarrow \mathbb{R}$ satisfies the following}:

\emph{(1). There exists $p>0, k>0$ such that for all $\boldsymbol{z} \in \mathcal{Z}$,}
\begin{equation}
0 \leq \mathcal{F}(\boldsymbol{z} ; \hat{\boldsymbol{y}}) \leq k\left(1+\|\boldsymbol{z}\|^p\right) .
\label{eq:cd1}
\end{equation}

\emph{(2). For every $c>0$ there is $k(c)>0$ such that for all $\boldsymbol{z},\ \boldsymbol{z}^{\prime} \in \mathcal{Z}$ with $\max \left\{\|\boldsymbol{z}\|,\left\|\boldsymbol{z}^{\prime}\right\|\right\}<c$,}
\begin{equation}
    \left|\mathcal{F}(\boldsymbol{z})-\mathcal{F}\left(\boldsymbol{z}^{\prime}\right)\right| \leq k(c)\left\|\boldsymbol{z}-\boldsymbol{z}^{\prime}\right\| .
    \label{eq:cd2}
\end{equation}
\emph{Proof.} 
As we use spectral normalization in the generator network $\boldsymbol{G}^*$, we can conclude that  $\boldsymbol{G}^{*}$ is $L$-Lipschitz from~\citet{miyato2018spectral} (see Section 2.1). 

(1) Since $\left\|\hat{\boldsymbol{y}}-\boldsymbol{A} \boldsymbol{G}^{*}(\boldsymbol{z})\right\|^2 \leq \left\|\hat{\boldsymbol{y}}\right\|^2 + \left\|\boldsymbol{A} \boldsymbol{G}^{*}(\boldsymbol{z})\right\|^2 \leq 2\left\|\hat{\boldsymbol{y}}\right\|^2 + 2\left\|\boldsymbol{A} \boldsymbol{G}^{*}(\boldsymbol{z})\right\|^2$, we can obtain directly that
\begin{equation}
\nonumber
\mathcal{F}(\boldsymbol{z} ; \hat{\boldsymbol{y}}) =\frac{\left\|\hat{\boldsymbol{y}}-\boldsymbol{A} \boldsymbol{G}^{*}(\boldsymbol{z})\right\|^2}{2 \sigma^2} \leq \frac{1}{\sigma^2}\left(\|\hat{\boldsymbol{y}}\|^2+\left\|\boldsymbol{A} \boldsymbol{G}^{*}(\boldsymbol{z})\right\|^2\right).
\end{equation}
Let $\boldsymbol{z}$ be any element in $\mathcal{Z}$, and we have:
\begin{equation}
\nonumber
\|\boldsymbol{G}^{*}(\boldsymbol{z})-\boldsymbol{G}^{*}(0)\|^2 \leq L\|\boldsymbol{z}\|^2.
\end{equation}
It follows that
\begin{equation}
\nonumber
\begin{aligned}
       \|\boldsymbol{G}^{*}(\boldsymbol{z})\|^2 &\leq 2\langle \boldsymbol{G}^{*}(\boldsymbol{z}), \boldsymbol{G}^{*}(0) \rangle + L\|\boldsymbol{z}\|^2-\|\boldsymbol{G}^{*}(0)\|^2 \\
       &\leq \frac{1}{2}\|\boldsymbol{G}^{*}(\boldsymbol{z})\|^2 +2\|\boldsymbol{G}^{*}(0)\|^2+L\|\boldsymbol{z}\|^2-\|\boldsymbol{G}^{*}(0)\|^2 \\
       &=2\|\boldsymbol{G}^{*}(0)\|^2+2L\|\boldsymbol{z}\|^2 \\
       &\leq B^2(1+\|\boldsymbol{z}\|^2),
\end{aligned}
\end{equation}
where $B$ denotes any constant that makes the above inequality hold.
Therefore, the function $\mathcal{F}$ satisfies
\begin{equation}
\nonumber
\begin{aligned}
\mathcal{F}(\boldsymbol{z} ; \hat{\boldsymbol{y}}) &\leq \frac{1}{\sigma^2}\left(\|\hat{\boldsymbol{y}}\|^2+\left\|\boldsymbol{A} \boldsymbol{G}^{*}(\boldsymbol{z})\right\|^2\right) 
\\
&\leq \frac{1}{\sigma^2}(\|\hat{\boldsymbol{y}}\|^2+\|\boldsymbol{A}\|^2B^2(1+\|\boldsymbol{z}\|^2)) \\
&\leq \frac{1}{\sigma^2} \max \{\|\hat{\boldsymbol{y}}\|^{2}, B^{2}\|\boldsymbol{A}\|^{2}\}\left (1+\|\boldsymbol{z}\|^2\right ),
\end{aligned}
\end{equation}
where $\frac{1}{\sigma^{2}} \max \{\|\hat{\boldsymbol{y}}\|^{2}, B^{2}\|\boldsymbol{A}\|^{2}\}$ corresponds to $k$ in \refeq{eq:cd1}.

(2) For every $c>0$,  if $\max \left\{\|\boldsymbol{z}\|,\left\|\boldsymbol{z}^{\prime}\right\|\right\}<c$, we have 
\begin{equation}\nonumber
\begin{aligned}
\left\|\boldsymbol{G}^{*}(\boldsymbol{z})+\boldsymbol{G}^{*}\left(\boldsymbol{z}^{\prime}\right)\right\| &\leq 2\|\boldsymbol{G}^{*}(0)\|+L\|\boldsymbol{z}\|+L\left\|\boldsymbol{z}^{\prime}\right\| 
\\
&\leq 2(\|\boldsymbol{G}^{*}(0)\|+L c).
\end{aligned}
\end{equation}
Once again we apply the $L$-Lipschitz property of $\boldsymbol{G}^{*}$, and by some elementary calculations we can derive
\begin{equation}
\nonumber
\begin{aligned}
\left|\mathcal{F}(\boldsymbol{z})-\mathcal{F}\left(\boldsymbol{z}^{\prime}\right)\right| &=\frac{1}{2 \sigma^2}\left | \left\|\hat{\boldsymbol{y}}-\boldsymbol{A} \boldsymbol{G}^{*}(\boldsymbol{z})\right\|^2-\left\|\hat{\boldsymbol{y}}-\boldsymbol{A} \boldsymbol{G}^{*}\left(\boldsymbol{z}^{\prime}\right)\right\|^2 \right | \\
&=\frac{1}{2 \sigma^2} \left | \|\boldsymbol{A} \boldsymbol{G}^{*}(\boldsymbol{z})\|^2-\|\boldsymbol{A} \boldsymbol{G}^{*}(\boldsymbol{z}^{\prime})\|^2 + 2 \hat{\boldsymbol{y}}^{\dagger} \boldsymbol{A}(\boldsymbol{G}^{*}(\boldsymbol{z})-\boldsymbol{G}^{*}(\boldsymbol{z}^{\prime})) \right|\\
&\leq \frac{1}{2 \sigma^2}\left | 2 \boldsymbol{A}L\hat{\boldsymbol{y}}^{\dagger} \|\boldsymbol{z}-\boldsymbol{z}^{\prime}\|+2\|\boldsymbol{A}\|^2 \|\boldsymbol{G}^{*}(\boldsymbol{z})-\boldsymbol{G}^{*}(\boldsymbol{z}^{\prime})\| \| \boldsymbol{G}^{*}(\boldsymbol{z})+\boldsymbol{G}^{*}(\boldsymbol{z}^{\prime}) \| \right |\\
&\leq \frac{1}{2 \sigma^2} \left \lvert 2\boldsymbol{A}^2L\|\boldsymbol{z}-\boldsymbol{z}^{\prime}\| (\|\boldsymbol{G}^{*}(0)\|+Lc)+2 \boldsymbol{A}L\hat{\boldsymbol{y}}^{\dagger} \|\boldsymbol{z}-\boldsymbol{z}^{\prime}\| \right\rvert\\
& \leq \frac{L\|\boldsymbol{A}\|}{\sigma^2}(\lvert \hat{\boldsymbol{y}}\rvert+\|\boldsymbol{A}\|(\|\boldsymbol{G}^{*}(0)\|+Lc))\|\boldsymbol{z}-\boldsymbol{z}^{\prime}\|, 
\end{aligned}
\end{equation}
where $(\lvert \hat{\boldsymbol{y}}\rvert+\|\boldsymbol{A}\|(\|\boldsymbol{G}^{*}(0)\|+Lc))$ corresponds to $k(c)$ in \refeq{eq:cd2}.

\subsection{The complete algorithm}
\begin{figure}[H]\centering
    \includegraphics[height = 0.62\textheight]{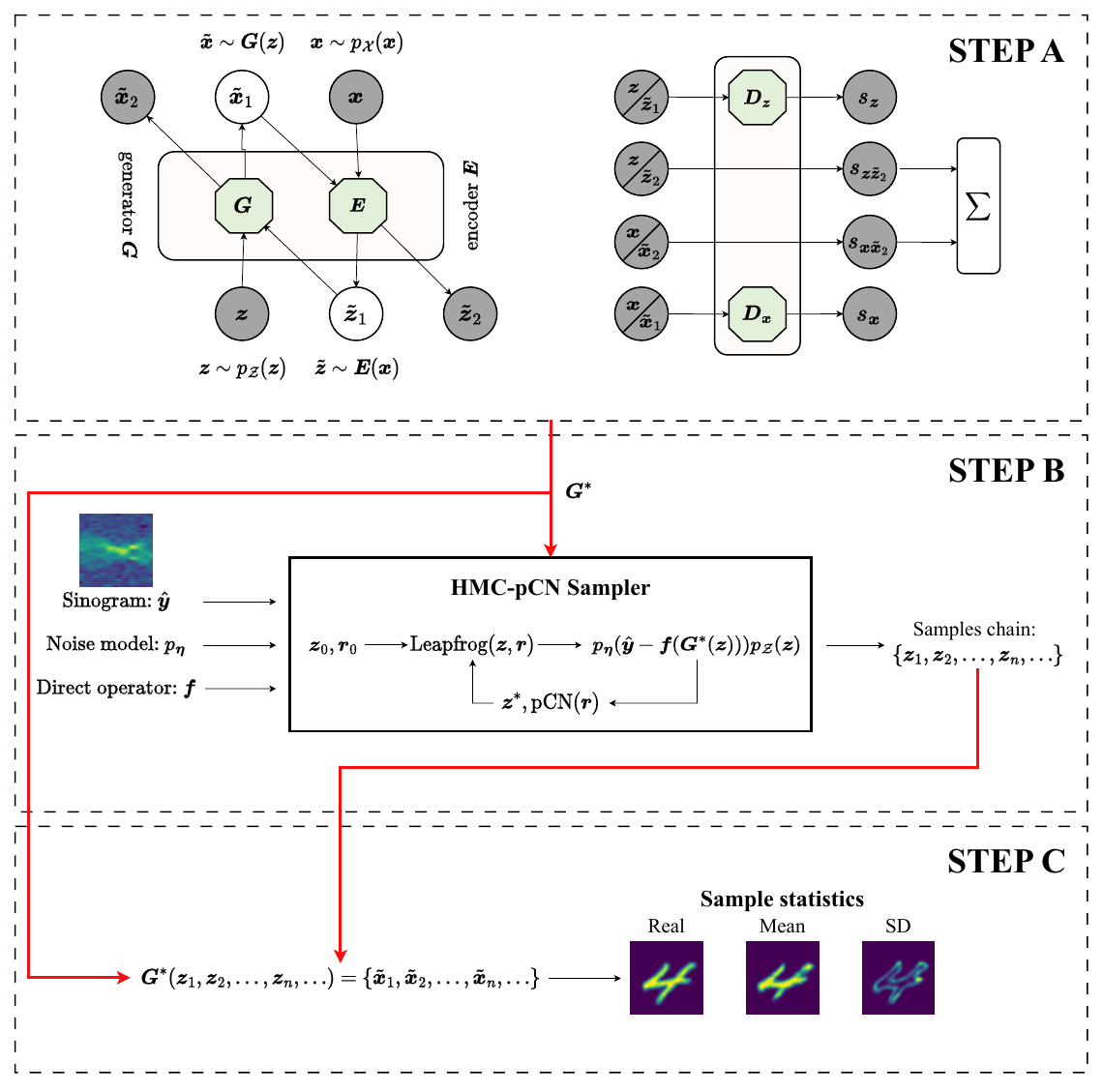}
    \caption{
        Overall inversion algorithm diagram. STEP A: Use existing samples to train SA-Roundtrip. STEP B: Generators, sinogram, direct operator, and noise distributions are applied to HMC-pCN sampler to obtain the posterior distribution in latent space. STEP C: Samples are taken from a posterior distribution of latent space using a convergent Markov chain, and the samples are generated by a trained generator to infer the field. 
    }
     \label{fig:full_alg}
\end{figure}

\reffig{fig:full_alg} shows the proposed Bayesian inference method with SA-Roundtrip prior, which involves three phases.

\begin{adjustwidth}{3.25em}{} 
    \begin{enumerate}
    \item[\textbf{Step A}] We train the proposed SA-Roundtrip prior using a large collection of available samples.
    \item[\textbf{Step B}] Based on the well-trained SA-Roundtrip prior of Step A, we generate samples from the posterior distribution in the latent space using the proposed HMC-pCN sampler.
    \item[\textbf{Step C}] We perform the point estimation as well as uncertainty quantification using the posterior samples obtained in Step B. 
    \end{enumerate}
\end{adjustwidth}

\section{Numerical Experiments and Result}\label{sec:exp}
% problem setup
In this section, we conduct experiments on the datasets of MNIST and TomoPhantom with two different dimensions to illustrate the performance of our proposed method.

We make the direct operator $\boldsymbol{f}$ Radon transform, which is routinely used in computerized tomography (CT)~\citep{Natterer1986}. And the number of projection angles of the Radon transform is always $\frac{\sqrt{m_1\times m_2}}{2}$, where $m_1 \times m_2$ is the shape of the samples and usually $m_1 = m_2$. The code and data utilized are available at 
\href{https://github.com/qjy415417122/SA-Roundtrip}{https://github.com/qjy415-417122/SA-Roundtrip}.
\subsection{Model Architecture}\label{sec:model_arch}

Compared with the traditional GAN, we replace convolution layers and fully connected layers (except the output layer and input layer) with residual blocks, and in each block, the normalization layer uses filter response normalization and the activation layer uses threshold linear unit (except for $\boldsymbol{D_z}$ and $\boldsymbol{D_x}$). Attention blocks were added to $\boldsymbol{G}$ and $\boldsymbol{D_x}$ when the resolution was half that of the target image to increase the model's capacity to represent global structure. In addition, the Spectral Normalisation is only used in $\boldsymbol{G}$ and $\boldsymbol{D_x}$. 

Encoder $\boldsymbol{E}$ and $\boldsymbol{D_x}$ are similar, but attention blocks are not used in $\boldsymbol{E}$. Similarly, the discriminator $\boldsymbol{D_z}$ does not use any normalized layer or attention block, and the convolution layer in the residual block is completely replaced by the dense layer. A more detailed model architecture is shown in \autoref{tb:arc_detail} and we have just adjusted the model depth for datasets with various levels of resolution. 
\begin{table}[!htb]
\caption{\\Details of SA-Roundtrip architecture for $32 \times 32$ resolution. The last item of input and output size is the feature dimension. $ConvT$ : transposed convolutional layer, $Conv$ : convolutional layer, $SN$ : spectral normalization, $FC$ : fully connected layer, $BN$ : batch normalization, $FRN$ : filter response normalization, $TLU$: threshold linear unit. $Pool$ denotes average pooling and $Global\ Pool$ denotes global average pooling. $Resblock-up$, $Resblock-down$, and $Resblock$ represent three residual blocks with up-sampling, down-sampling, and invariant resolution, respectively.}
\centering
\resizebox{\textwidth}{!}{
\begin{tabular}{ccccc}
\toprule
\hline  & Generator & Discriminator\_x & Encoder & Discriminator\_z \\
\hline Input size & $1 \times 1 \times 24$ & $32 \times 32 \times 1$ & $32 \times 32 \times 1$ & $1 \times 1 \times 24$ \\
 {$\begin{array}{c} \\ \text{Stage 1} \\ \\\end{array}$} 
& {$\begin{array}{c} \text{ConvT+SN}\end{array}$} 
& {$\begin{array}{c} \text{Conv+SN+} \\ \text{LeakyReLU+Pool}  \end{array}$}
& {$\begin{array}{c} \text{Conv+FRN+} \\ \text{TLU+Pool} \end{array}$} 
& {$\begin{array}{c} \text{Flatten+FC} \\ \text{+Resblock}  \end{array}$} \\
 Output size & $4 \times 4 \times 256$ & $16 \times 16 \times 64$ & $16 \times 16 \times 64$ & $1\times256$ \\
 {$\begin{array}{c} \\ \text{Stage 2} \\ \\\end{array}$} 
& {$\begin{array}{c} \text{Resblock-up} \times 2 \\ \text{+SA-block}\end{array}$} 
& {$\begin{array}{c} \text{SA-block+}\\ \text{Resblock-down}\times 2 \end{array}$}
& {$\begin{array}{c}\text{Resblock-down} \times 2\end{array}$} 
& {$\begin{array}{c} \text{Resblock} \times 2 \end{array}$} \\
 Output size & $16 \times 16 \times 64$ & $4 \times 4 \times 256$ & $4 \times 4 \times 256$ & $1\times256$ \\
 {$\begin{array}{c} \\ \text{Stage 3} \\ \\ \end{array}$} 
& {$\begin{array}{c}\text{Resblock-up+}\\ \text{FRN+TLU+ConvT}\end{array} $} 
& {$\begin{array}{c}\text{LeakyReLU+} \\ \text{Global Pool+FC} \end{array}$} 
& {$\begin{array}{c} \text{FRN+TLU}\\ \text{+Conv} \end{array}$} 
& {$\begin{array}{c} \text{BN+Tanh}\\ \text{+FC}\end{array}$} \\
 Output size & $32 \times 32 \times 1$ & $1 \times 1$ & $1 \times 1 \times 24$ & $1 \times 1$ \\
\hline
\bottomrule
\end{tabular}}\label{tb:arc_detail}
\end{table} 
        
\subsection{Generative capability of SA-Roundtrip}
\begin{figure}[!htb]
    \centering
    \subfloat[The true samples $\boldsymbol{x}$]{\includegraphics[height = 0.3\textheight]{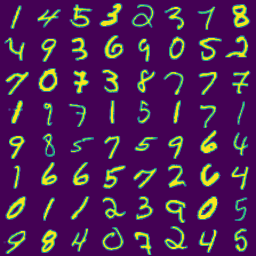}%
        \label{fig2_first_case}}
        \hfil
    \subfloat[The generated samples $\tilde{\boldsymbol{x}}$]{\includegraphics[height = 0.3\textheight]{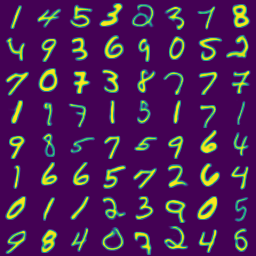}%
            \label{fig2_second_case}}
    \caption{
    MNIST dataset: (a) Test samples $\boldsymbol{x}$. (b) Generated reconstructions   $\tilde{\boldsymbol{x}}=\boldsymbol{G}^*(\boldsymbol{E}(\boldsymbol{x}))$. 
    }
    \label{fig:2} 
\end{figure}

\begin{figure}[!htb]
    \centering
    \subfloat[The true samples $\boldsymbol{x}$]{\includegraphics[height = 0.3\textheight]{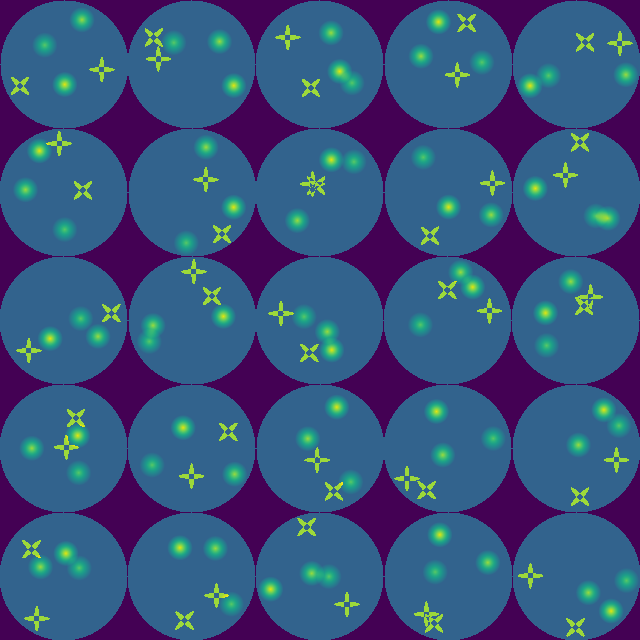}%
        \label{fig3_first_case}}
        \hfil
    \subfloat[The generated samples $\tilde{\boldsymbol{x}}$]{\includegraphics[height = 0.3\textheight]{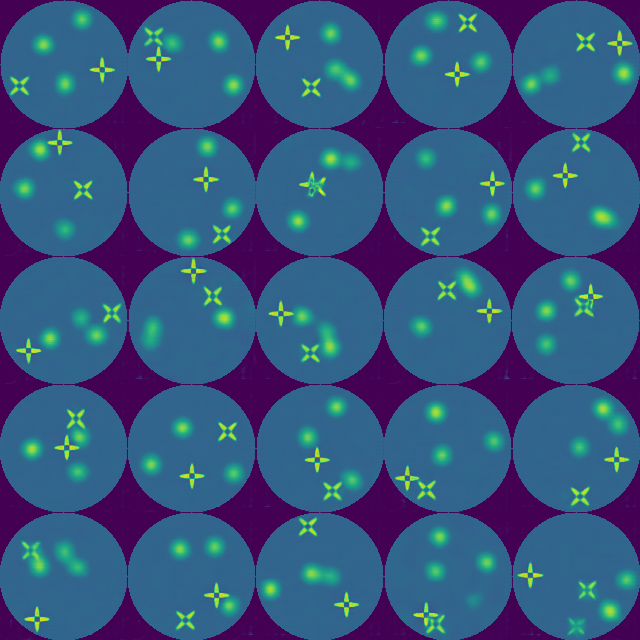}%
            \label{fig3_second_case}}
    \caption{
    TomoPhantom dataset: (a) Test samples $\boldsymbol{x}$. (b) Generated reconstructions   $\tilde{\boldsymbol{x}}=\boldsymbol{G}^*(\boldsymbol{E}(\boldsymbol{x}))$. 
    }
    \label{fig:3}
\end{figure}

\textcolor{black}{An essential criterion for evaluating the effectiveness of the SA-Roundtrip prior is its ability to generate samples that closely resemble real data samples.}
The model should generate a sample $\tilde{\boldsymbol{x}} = \boldsymbol{G}^*(\boldsymbol{E}(\boldsymbol{x}))$ that is sufficiently similar to the ground truth sample $\boldsymbol{x}$.
% The quality of this metric may directly affect the reconstruction effect of Bayesian inference. 
Figs. \ref{fig:2} and \ref{fig:3} show the samples and their corresponding generated samples for the MNIST and TomoPhantom datasets, respectively. It is evident that the vast majority of the generated samples are almost identical to the real samples. 

\subsection{Reconstruct using Bayesian inference}

\subsubsection{Identify potential dimensions}
The structure of encoder networks $\boldsymbol{E}$ makes it possible to observe the dimension of the latent space. 
In order to accurately approach the true distribution, the dimensions of the latent space should at least be equal to the potential dimensions of the data. In addition, overestimating dimensions results in less regularization (and non-convergence problems due to increased computational complexity). 

\begin{figure}[!htb]
    \centering
    \subfloat[]{\includegraphics[height=6.5cm]{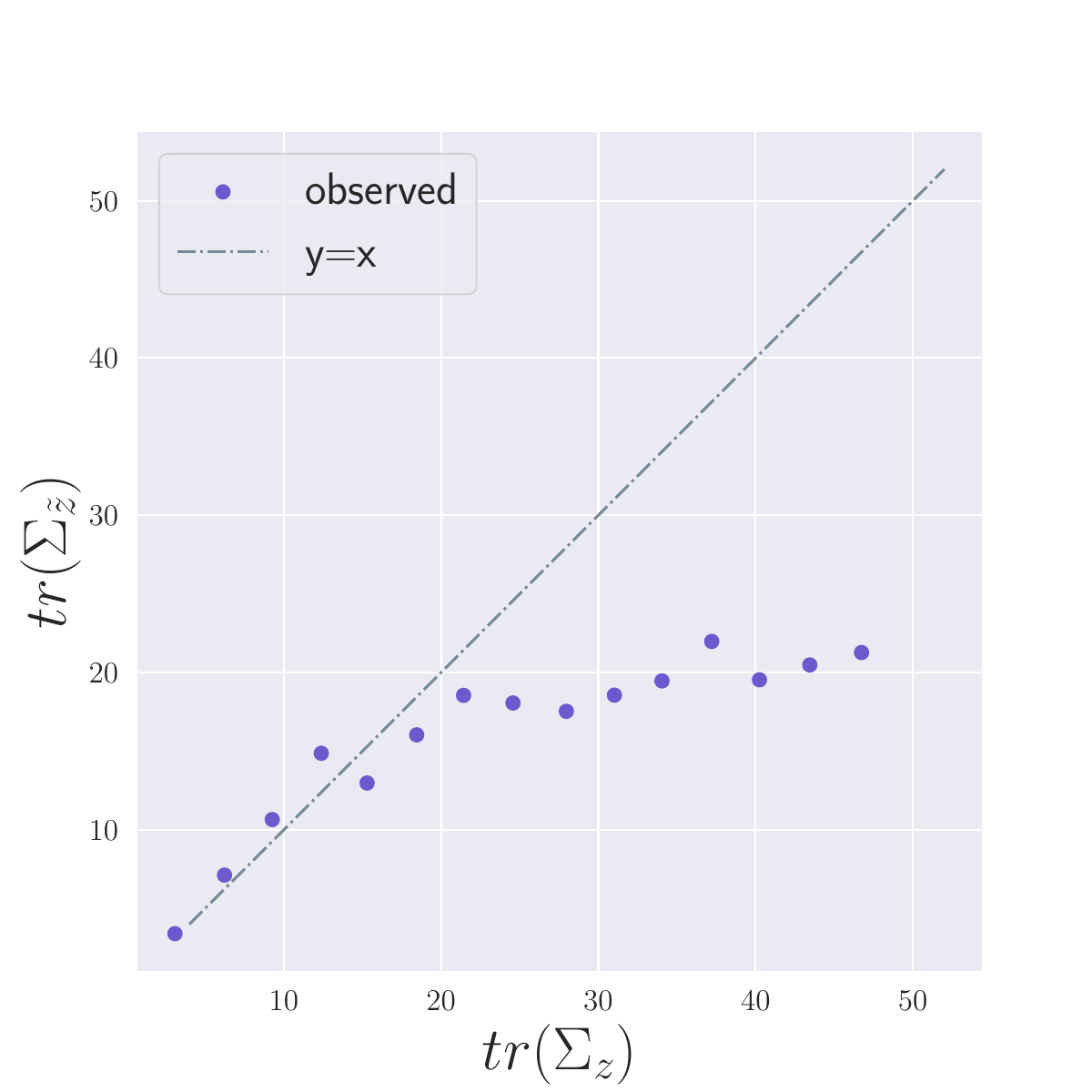} \label{fig:dim_ana(a)}} \quad
    \subfloat[]{\includegraphics[height=5.7cm]{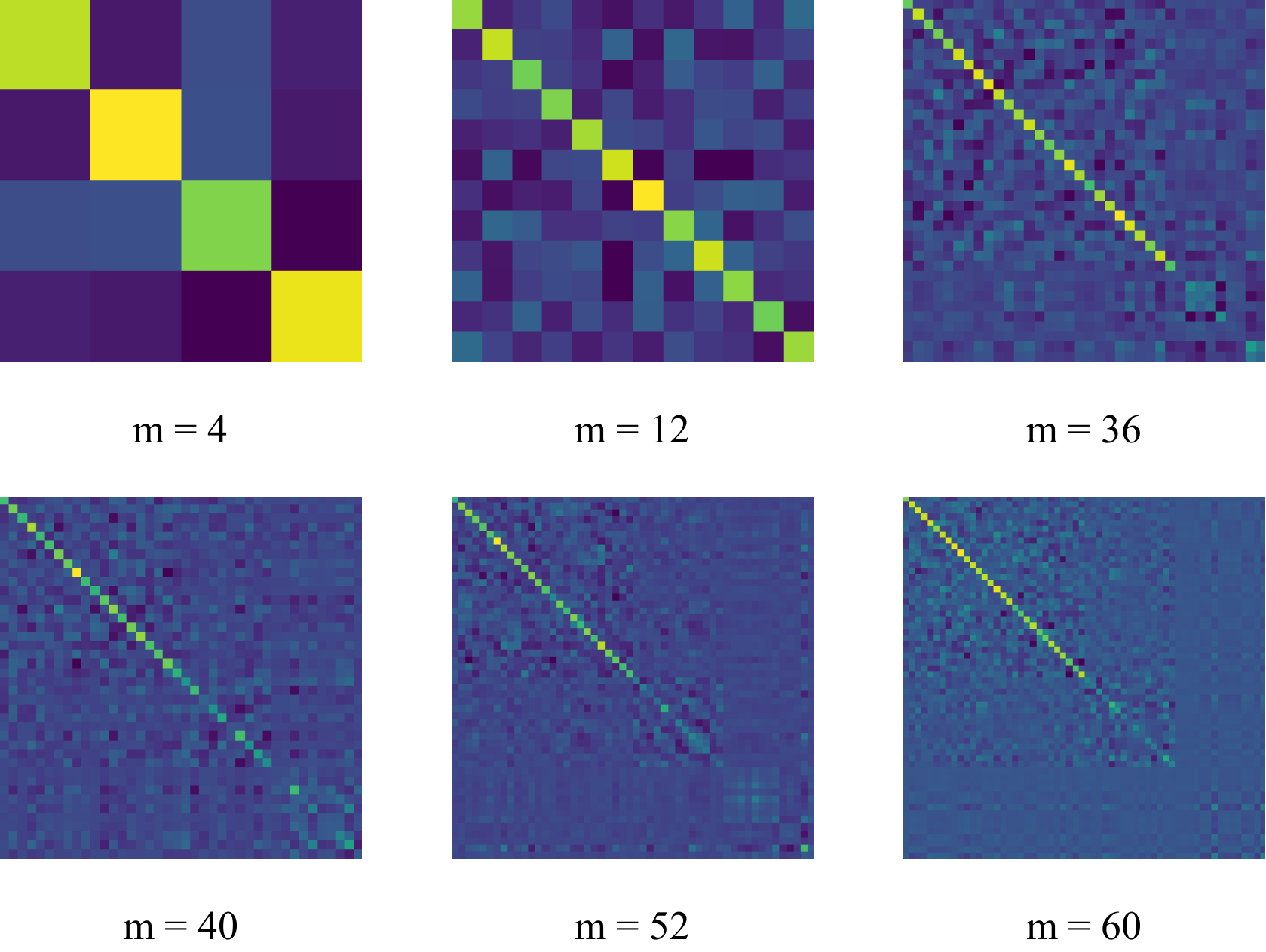} \label{fig:dim_ana(b)}}
    \caption{Dimensional analysis. (a) Trace of  $\Sigma_{\tilde{\boldsymbol{z}}}$ against trace of $\Sigma_{\boldsymbol{z}}$. (b) The sample covariance matrix $\Sigma_{\tilde{\boldsymbol{z}}}$. The dimension of $\tilde{\boldsymbol{z}}$ is $m$. The greater brightness along the diagonal of the image indicates the higher significance of the corresponding dimension.}
    \label{fig:dim_ana}
\end{figure}

This can be evaluated by calculating the trace of the covariance of the sample generated by the encoder \citep{holden2021bayesian}. 
\textcolor{black}{
When SA-Roundtrip accurately captures the underlying manifold of $\{\boldsymbol{x}_i\}^n_{i=1}$, we anticipate that $\{\tilde{\boldsymbol{z}}_i\}^n_{i=1}$ will contain information about the encoding uncertainty. We first compute the trace of covariance matrix of $\{\tilde{\boldsymbol{z}}_i\}^n_{i=1}$: 
\begin{equation*}
    tr(\Sigma_{\tilde{\boldsymbol{z}}}) = \sum\limits_{i}^{m}(\Sigma_{\tilde{\boldsymbol{z}}})_{ii},
\end{equation*}
where
\begin{equation}\nonumber
    \begin{array}{cc}
             &\Sigma_{\tilde{\boldsymbol{z}}} = \frac{1}{n-1}\sum (\tilde{\boldsymbol{z}}_i-\bar{\boldsymbol{z}})^T (\tilde{\boldsymbol{z}}_i-\bar{\boldsymbol{z}}).
    \end{array}
\end{equation}
We then compare it with the trace of covariance, $tr(\Sigma_{\boldsymbol{z}})$, for the set $\{\boldsymbol{z}_i\}^n_{i=1}$. 
If $tr(\Sigma_{\boldsymbol{z}}) > tr(\Sigma_{\tilde{\boldsymbol{z}}})$, it indicates an excess of dimensions, specifically, that the dimensionality of the latent space surpasses the intrinsic dimensionality of the manifold supporting $\{\boldsymbol{x}_i\}^n_{i=1}$.} 
In the context of MNIST, as shown in \reffig{fig:dim_ana(a)}, the case occurs for dimensions exceeding 24. Therefore, we choose that the dimension of the latent space be 24. Furthermore, \reffig{fig:dim_ana(b)} depicts the covariance matrices of $\{\tilde{\boldsymbol{z}}_i\}^n_{i=1}$, with their traces corresponding to the points shown in \reffig{fig:dim_ana(a)}.

\begin{figure}[!htb]
    \centering
    \includegraphics[height = 0.3\textheight]{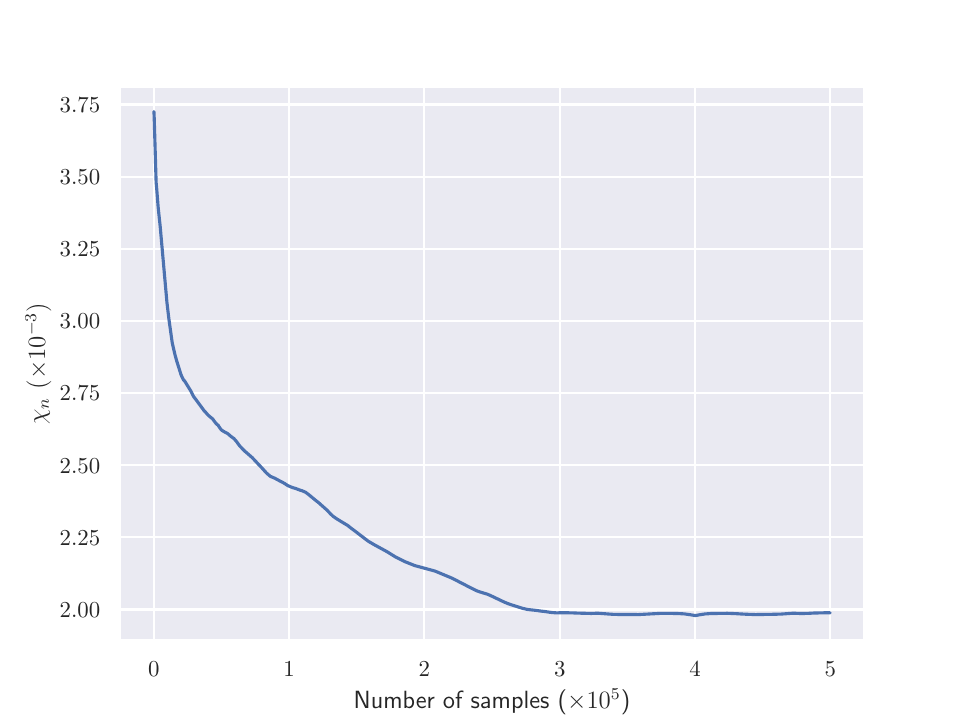}
    \caption{\textcolor{black}{The discrepancy $\chi_n$ with respect to different sample sizes.}
    }
    \label{fig:post_con}
\end{figure}

\textcolor{black}{\subsubsection{Convergence of posterior}
In order to demonstrate the convergence of posterior statistics obtained through latent space sampling, we compute the difference of the posterior mean in $\boldsymbol{x}$ and the posterior mean in $\boldsymbol{z}$ space. We use a total variation prior, and then we rewrite the posterior distribution \refeq{eq:bayes_infer} as:
\begin{equation}\nonumber
\begin{aligned}
p_{\mathcal{X}}^{\text {post }}(\boldsymbol{x} \mid \hat{\boldsymbol{y}}) 
&\propto p_{\boldsymbol{\eta}}(\hat{\boldsymbol{y}}-\boldsymbol{f}(\boldsymbol{x})) p_{\mathcal{X}}(\boldsymbol{x})\\
&\propto \exp \left\{-\frac{\left(\hat{\boldsymbol{y}}-\boldsymbol{Ax}\right)^{\dagger}\left(\hat{\boldsymbol{y}}-\boldsymbol{Ax}\right)}{2 \sigma^2}-\tau\|\boldsymbol{x}\|_{TV}\right\},
\end{aligned}
\end{equation}
where $\tau$ is a predefined positive constant, and $\|\boldsymbol{x}\|_{TV}$ denotes the total variation of $\boldsymbol{x}$. 
We set $\tau=10$ for optimal performance and sample from $p_{\mathcal{X}}^{\text {post }}(\boldsymbol{x} \mid \hat{\boldsymbol{y}})$ using the pCN algorithm. 
Subsequently, we compute the mean squared error of the posterior samples $\{\boldsymbol{x}_i\}_{i=1}^{n}$ from $p_{\mathcal{X}}^{\text {post }}(\boldsymbol{x} \mid \hat{\boldsymbol{y}})$ and the posterior samples $\{\boldsymbol{G}^*(\boldsymbol{z}_i)\}_{i=1}^{n}$ from $p_{\mathcal{Z}}^{\text {post }}(\boldsymbol{z} \mid \hat{\boldsymbol{y}})$ in Eq. \eqref{eq:post_dis}. This discrepancy is defined as:
\begin{equation}\nonumber
    \chi_n = \|\overline{\boldsymbol{x}}_n-\overline{\boldsymbol{G}^*(\boldsymbol{z}_n)}\|_2^2,
\end{equation}
where $\overline{\boldsymbol{x}}_n = \frac{1}{n} \sum\limits_{i}^{n} \boldsymbol{x}_i$ and $\overline{\boldsymbol{G}^*(\boldsymbol{z}_n)} = \frac{1}{n} \sum\limits_{i}^{n} \boldsymbol{G}^*(\boldsymbol{z}_i)$. \reffig{fig:post_con} shows a small and stabilized difference when the sample size exceeds $3\times 10^5$, suggesting the convergence of the proposed approach in the latent space.}
 
\subsubsection{Performance comparison}
\begin{figure}[!htb]\centering
    \begin{tabular}{cc}

    \shortstack{(a) \\ \\ \\ \\ \\ \\ \\ \text{}} &
    \includegraphics[width=\reconwidth]{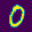}
     \includegraphics[width=\reconwidth]{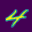}
     \includegraphics[width=\reconwidth]{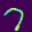}
     \includegraphics[width=\reconwidth]{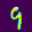}
    \\
    \shortstack{(b) \\ \\ \\ \\ \\ \\ \\ \text{}} &
    \includegraphics[width=\reconwidth]{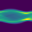}
     \includegraphics[width=\reconwidth]{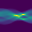}
     \includegraphics[width=\reconwidth]{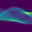}
     \includegraphics[width=\reconwidth]{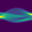}
    \\
     \shortstack{(c) \\ \\ \\ \\ \\ \\ \\ \text{}} &
     \includegraphics[width=\reconwidth]{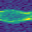}
     \includegraphics[width=\reconwidth]{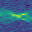}
     \includegraphics[width=\reconwidth]{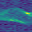}
     \includegraphics[width=\reconwidth]{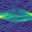}
    \end{tabular}
     \\
     \caption{True (test) samples from MNIST and the corresponding sinograms. (a) Sample of the ground truth. (b) The corresponding clean sinogram. (c) The corresponding sinogram with 10\% noise levels.}
     \label{fig:5}
\end{figure}

In this study, we compare the proposed method with the filtered back projection (FBP) and WGAN-GP~\citep{gulrajani2017improved} approaches.
We conduct separate experiments using two datasets. \textcolor{black}{Note that we train the SA-Roundtrip model using clean data. The first experiment uses the MNIST dataset, which includes 60,000 handwritten digit images, each at 32$\times$32 pixels resolution.} The resulting image from direct operator $\boldsymbol{f}$ is known as a sinogram, and \reffig{fig:5}$(\mathrm{b})$ shows some clean examples. We choose some test samples from the true prior (see \reffig{fig:5}$(\mathrm{a})$), and add Gaussian noise $\boldsymbol{\eta} \sim \mathcal{N}\left(0, \sigma^{2} \mathbb{I}_{p}\right)$ to the sinogram to generate the noisy measurement $\hat{\boldsymbol{y}}$ (see \reffig{fig:5}$(\mathrm{c})$). 
We set signal noise ratio $\gamma$ to 0.1 and $\sigma = \max (\lvert \hat{\boldsymbol{y}} \rvert)$ (maximum of the absolute value of sinogram matrix). The value of $\sigma^{2}$ is usually greater than 20, which will be a destructive noise item.
\begin{figure}[H]
    \centering
    \includegraphics[height = 0.4\textheight]{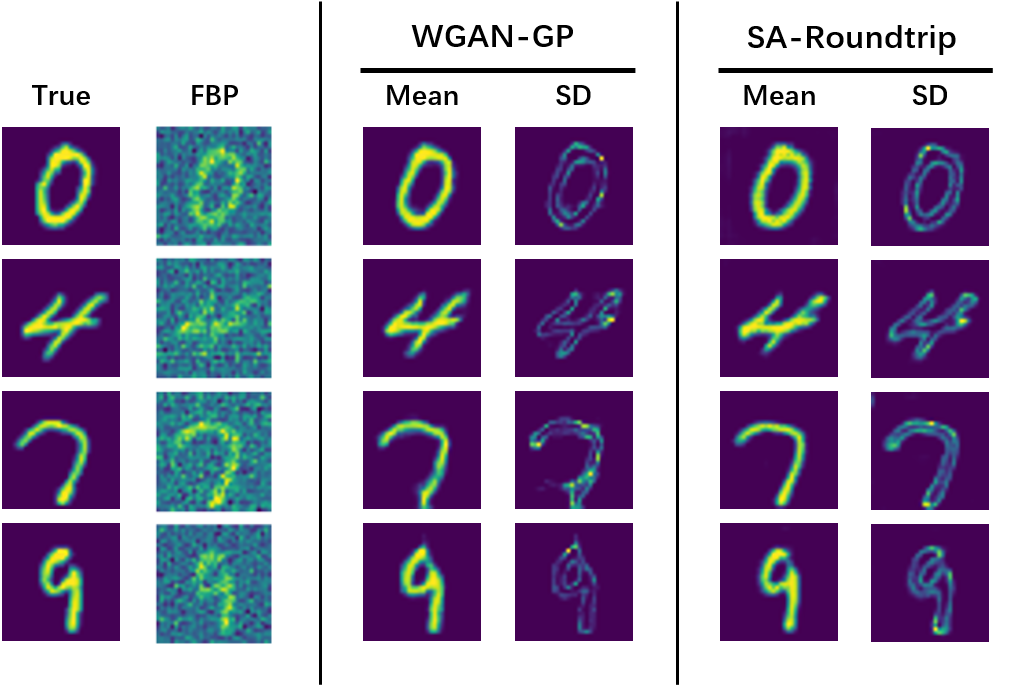}
    \begin{tabular}{lllll}
         &  &  &  & \\
    \end{tabular}
    \resizebox{\textwidth}{!}{
    \begin{tabular}{ccccc@{\hspace{25pt}}cccc@{\hspace{25pt}}}
    \toprule
        \hline & \multicolumn{4}{c}{ PSNR }\hspace{25pt} & \multicolumn{4}{c}{ SSIM }\hspace{25pt} \\
        \hline \multicolumn{1}{c}{ Samples } & Digit\ 0 & Digit\ 4 & Digit\ 7 & Digit\ 9 & Digit\ 0 & Digit\ 4 & Digit\ 7 & Digit\ 9 \\
         \multicolumn{1}{c}{ FBP } & $12. 07$ & $9. 47$ & $11. 77$ & $11. 46$ & $0.53$ & $0.32$ & $0.37$ & $0.37$ \\
        \multicolumn{1}{c}{ WGAN-GP } & $25. 35$ & $24. 21$ & $23. 23$ & $23. 85$ & \textcolor{black}{0.93} & $0.87$ & $0.89$ & $0.89$ \\
        \multicolumn{1}{c}{ SA-Roundtrip } & \textcolor{black}{27.46} & \textcolor{black}{25.72} & \textcolor{black}{28.94} & \textcolor{black}{28.31} & $0.92$ & \textcolor{black}{0.88} & \textcolor{black}{0.91} & \textcolor{black}{0.95} \\
        \hline
        \bottomrule
        \end{tabular}}
    \caption{
        Comparison of experimental results. \textcolor{black}{True: Sample of the ground truth. FBP: Results of FBP. Mean: $\sum_{i=1}^{n}\boldsymbol{G}^*(\boldsymbol{z}_i)/n$. SD: standard deviation of $\{\boldsymbol{G}^*(\boldsymbol{z}_i)\}_{i=1}^n$.} It can be clearly seen that the results of the deep generative prior at destructive noise levels have significant advantages over those of FBP. Moreover, the results of SA-Roundtrip are also better than those of WGAN-GP.
    }
    \label{fig:6}
\end{figure}
\begin{table}[!htb]
\centering
\caption{\\ \textcolor{black}{The PSNR and SSIM values for different data sizes. The \emph{Dataset Ratio} indicates the proportion of the training dataset used. The best results are indicated by red color.}}
\begin{tabular}{cccc}
\toprule
\hline
Dataset Ratio & SSIM & PSNR \\
\midrule
$25\%$ & 0.86 & 22.42 \\
$50\%$ & 0.88 & 26.36 \\
$75\%$ & 0.91 & 27.01 \\
$100\%$ & \textcolor{black}{0.92} & \textcolor{black}{27.46} \\
\hline
\bottomrule
\end{tabular}\label{tb:model}
\end{table}
\begin{table}[!htb]
\centering
\caption{\textcolor{black}{Percentage of the real data pixels that fall within the estimated HPDI for 95\% and 99\% confidence level. A higher percentage indicates a more accurate estimation of uncertainty.}}
\begin{tabular}{ccc}
\toprule
\hline
Method & Confidence level & Percentage \\
\midrule
\multirow{2}{*}{SA-Roundtrip} & 95\% & {78.1\%} \\
& 99\% & {85.9\%} \\
\multirow{2}{*}{WGAN-GP} & 95\% & 75.3\% \\
& 99\% & 83.5\% \\
\hline
\bottomrule
\end{tabular}\label{tb:pe}
\end{table}

\textcolor{black}{
To analyse the data efficiency of the SA-Roundtrip prior, we compare the recovered test images using priors trained on different sample sizes. Specifically, we partitioned the training sets of the MNIST dataset into  25\%, 50\%, 75\%, and 100\% of the images, respectively. \autoref{tb:model} shows the mean peak signal-to-noise ratio (PSNR) and structural similarity index (SSIM) values of 10 reconstructed samples from the testing images. It is evident that a larger dataset results in improved performance.
}

Obviously, for more destructive noise levels, FBP performance will deteriorate dramatically(see \reffig{fig:6}). On the other hand, the proposed SA-Roundtrip model is more robust to highly challenging noise levels than WGAN-GP and FBP, highlighting the advantages of our model and Bayesian framework in addressing complex uncertainty problems.

\textcolor{black}{
Furthermore, it is important to assess the reliability of the uncertainty in a full Bayesian method.
This can be accomplished by computing the highest posterior density intervals (HPDI) and then calculating how many pixels of the ground truth image fit within this HPDI. For any confidence level $\alpha \in [0, 1]$, the 100(1 - $\alpha$)\% HPDI is defined as follows \citet{pereyra2017maximum}:
\begin{equation}
    \mathcal{I}_\alpha=\{\boldsymbol{G}^*(\boldsymbol{z})|p^{\text{post}}_{\mathcal{Z}}(\boldsymbol{z}|\hat{\boldsymbol{y}}) > p_\alpha\},
\end{equation}
where $p_\alpha$ is the largest constant satisfying $\mathbb{P}[\boldsymbol{G}^*(\boldsymbol{z})|p_{\mathcal{Z}}(\boldsymbol{z}|\hat{\boldsymbol{y}})>p_\alpha] = 1-\alpha$.
As demonstrated in \autoref{tb:pe}, SA-Roundtrip have significantly higher percentages values for both 95\% and 99\% confidence intervals, indicating better accuracy.
}

\begin{figure}[!htb]\centering
    \begin{tabular}{cccc}
    \shortstack{(a) \\ \\ \\ \\ \\ \\ \\ \text{}} &
    \includegraphics[width=\reconwidth]{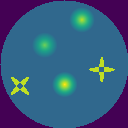}
     \includegraphics[width=\reconwidth]{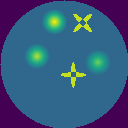}
     \includegraphics[width=\reconwidth]{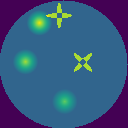}
     \includegraphics[width=\reconwidth]{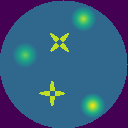}& & 
    \\
    \shortstack{(b) \\ \\ \\ \\ \\ \\ \\ \text{}} &
    \includegraphics[width=\reconwidth]{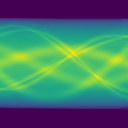}
     \includegraphics[width=\reconwidth]{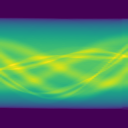}
     \includegraphics[width=\reconwidth]{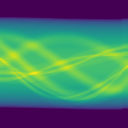}
     \includegraphics[width=\reconwidth]{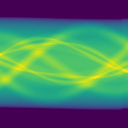}& & 
    \\
     \shortstack{(c) \\ \\ \\ \\ \\ \\ \\ \text{}} &
     \includegraphics[width=\reconwidth]{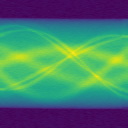}
     \includegraphics[width=\reconwidth]{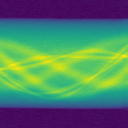}
     \includegraphics[width=\reconwidth]{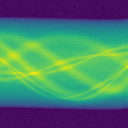}
     \includegraphics[width=\reconwidth]{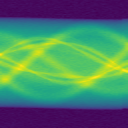}& & 
    \end{tabular}
     \\
     \caption{True (test) samples from TomoPhantom and the corresponding sinograms. (a) Sample of the ground truth. (b) The corresponding clean sinogram. (c) The corresponding sinogram with 1\% noise levels.}
     \label{fig:7}
    \end{figure}

\begin{figure}[H]
        \centering
        \includegraphics[height = 0.4\textheight]{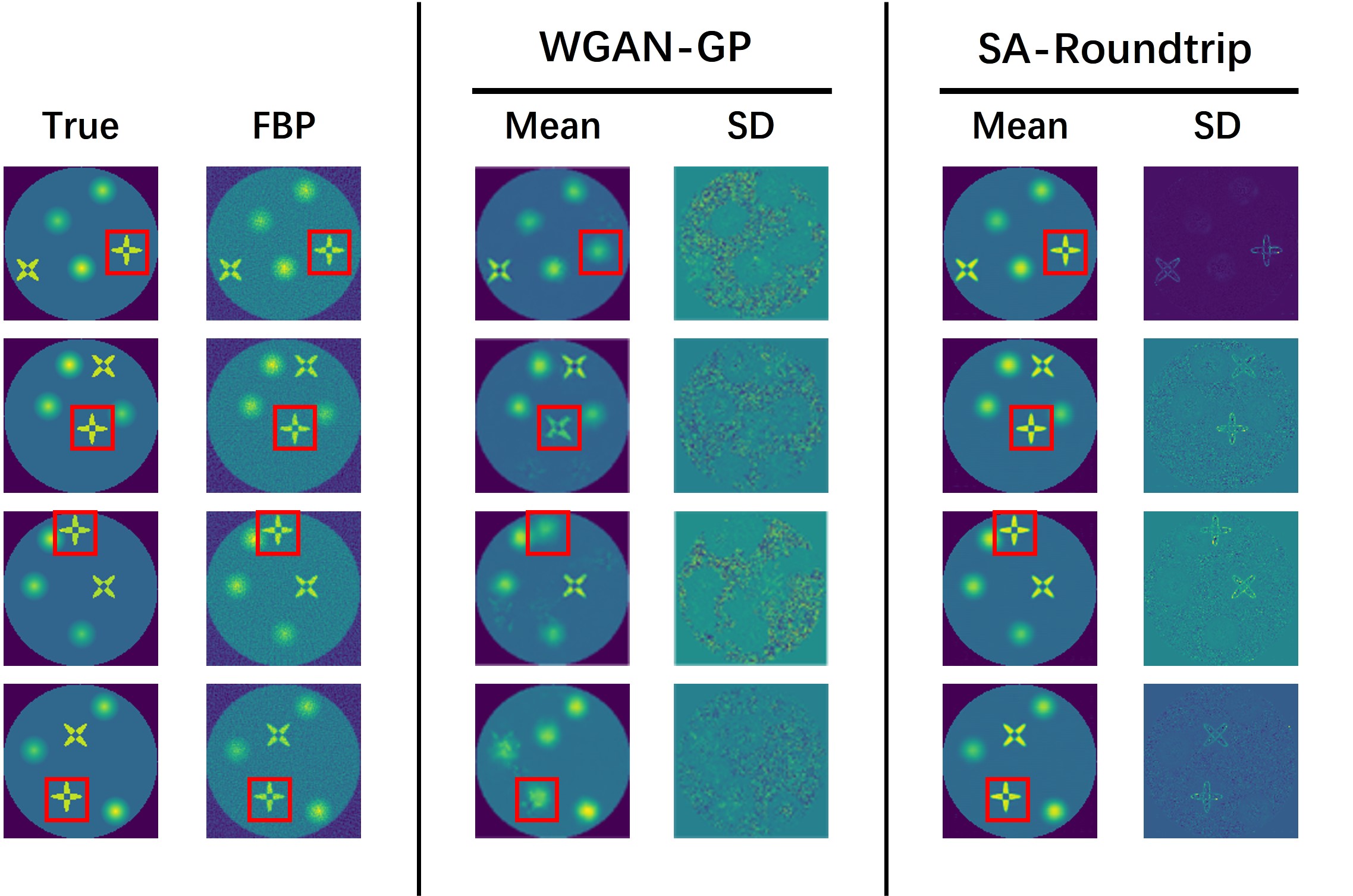}
        \begin{tabular}{lllll}
             &  &  &  & \\
        \end{tabular}
        \resizebox{\textwidth}{!}{
    \begin{tabular}{ccccc@{\hspace{25pt}}cccc@{\hspace{25pt}}}
    \toprule
        \hline & \multicolumn{4}{c}{ PSNR }\hspace{25pt} & \multicolumn{4}{c}{ SSIM }\hspace{25pt} \\
            \hline \multicolumn{1}{c}{ Samples } & Sample\ 0 & Sample\ 1 & Sample\ 2 & Sample\ 3 & Sample\ 0 & Sample\ 1 & Sample\ 2 & Sample\ 3 \\
             \multicolumn{1}{c}{ FBP } & $25. 84$ & $26. 11$ & $25. 92$ & $26. 02$ & $0.701$ & $0.717$ & $0.707$ & $0.713$ \\
            \multicolumn{1}{c}{ WGAN-GP } & $26. 855$ & $25. 538$ & $24. 740$ & $23. 568$ & $0.951$ & $0.935$ & $0.897$ & $0.875$ \\
            \multicolumn{1}{c}{ SA-Roundtrip } & \textcolor{black}{33.56} & \textcolor{black}{33.68} & \textcolor{black}{33.82} & \textcolor{black}{33.83} & \textcolor{black}{0.989} & \textcolor{black}{0.987} & \textcolor{black}{0.988} & \textcolor{black}{0.989} \\
            \hline
            \bottomrule
            \end{tabular}}
        \caption{Comparison of experimental results. From top to bottom are sample 0, sample 1, sample 2, and sample 3. \textcolor{black}{True: Sample of the ground truth. FBP: Results of FBP. Mean: $\sum_{i=1}^{n}\boldsymbol{G}^*(\boldsymbol{z}_i)/n$. SD: standard deviation of $\{\boldsymbol{G}^*(\boldsymbol{z}_i)\}_{i=1}^n$.} Obviously, in terms of metrics, better results are achieved by all three methods at \text{1\%} noise level and higher resolution images. Although WGAN-GP achieves similar results to FBP in PSNR, and even WGAN-GP has completely outperformed FBP in SSIM, we can see that WGAN-GP reconstructs certain features that are completely wrong \textcolor{black}{(that is, the location marked by the red squares in the figure)}. However, SA-Roundtrip achieves convincing results both in both metrics and in the actual reconstructed results.
        }
        \label{fig:8}
    \end{figure}

We then consider the same case as above, but with a different dataset and a different noise level, i.e. $\gamma=0.01$. We use TomoPhantom~\citep{KAZANTSEV2018150} to generate 40000 2D graphics with a resolution of 128 $\times$ 128, which are used for the benchmark test of CT image reconstruction. There are six different feature points: different color depths and different shapes. The locations of the feature points are sampled from a disc with a radius of 128 using a standard normal distribution. Some samples from the dataset and the sinograms for the experiment and their sinograms with noise are shown in \reffig{fig:7}$(\mathrm{a})$. The experimental data and results are shown in \reffig{fig:7} and \reffig{fig:8} respectively. It can be seen that the reconstruction effect of our model is much higher than that of FBP and WGAN-GP.
    
    \newpage
    
\section{Conclusion}\label{sec:conclusion}
Throughout this paper, we present a new Bayesian approach for solving linear imaging problems that is self-attention roundtrip (SA-Roundtrip) prior. 
This prior is based on self-attention generative adversarial networks 
and is able to effectively and robustly encode and decode bidirectionally. 
We then use the Hamiltonian Monte Carlo with pCN (HMC-pCN) to sample from the posterior distribution derived by the SA-Roundtrip prior, which has shown to be ergodic under particular assumptions. 
Finally, we illustrate the effectiveness of our method on computed tomography reconstruction tasks using the MNIST and TomoPhantom datasets, showing that the SA-Roundtrip prior outperforms state-of-the-art methods in inferring the uncertainty of reconstructions. 
We also presented results inferring the intrinsic dimension of MNIST data.
We believe that the SA-Roundtrip prior can be used to solve many other inverse problems, such as electrical impedance tomography and so on, which is our research interest in the future.

\nolinenumbers
\section*{Acknowledgements}\label{sec:acknowledgements}
The work is supported by the National Natural Science Foundation of China under Grant 12101614 and the Natural Science Foundation of Hunan Province, China, under Grant 2021JJ40715. We are grateful to the High-Performance Computing Center of Central South University for assistance with the computations.

\clearpage
\bibliographystyle{elsarticle-harv}
\bibliography{main-refs}

\end{document}